\documentclass[utf8]{FrontiersinHarvard}

\usepackage{url,hyperref,lineno,microtype} 
\usepackage[onehalfspacing]{setspace}

\usepackage{siunitx}
\usepackage{graphicx}
\usepackage{amsfonts,amssymb,amstext,amsmath,amsthm,verbatim,times,cancel,epsfig}
\usepackage{array}
\graphicspath{{figures/}}

\usepackage{vcell}
\usepackage{tabularx}
\usepackage{colortbl}
\usepackage{xcolor}

\usepackage{enumitem}
\usepackage{caption}
\usepackage{subfig}
\renewcommand{\thesubfigure}{\textbf{\Alph{subfigure}}}
\usepackage{soul}

\usepackage[export]{adjustbox}
\newcolumntype{P}[1]{>{\centering\arraybackslash}p{#1}}
\usepackage[colorinlistoftodos]{todonotes}

\def\keyFont{\fontsize{8}{11}\helveticabold}
\def\firstAuthorLast{Pace {et~al.}} 
\def\Authors{Anna Pace\,$^{1,*}$, Giorgio Grioli\,$^{1,2,3}$, Alice Ghezzi\,$^{1,2}$, Antonio Bicchi\,$^{1,2,3}$ and Manuel G. Catalano\,$^{1}$}
\def\Address{$^{1}$Soft Robotics for Human Cooperation and Rehabilitation Lab, Fondazione Istituto Italiano di Tecnologia, Genoa, Italy\\
$^{2}$Department of Information Engineering, University of Pisa, Pisa, Italy\\
$^{3}$Research center “E. Piaggio”, University of Pisa, Pisa, Italy }
\def\corrAuthor{Anna Pace}
\def\corrEmail{anna.pace@iit.it}

\begin{document}
\onecolumn
\firstpage{1}

\title[]{Investigating the Performance of Soft Robotic Adaptive Feet with Longitudinal and Transverse Arches}

\author[\firstAuthorLast ]{\Authors} 
\address{} 
\correspondance{} 

\extraAuth{}

\maketitle

\begin{abstract}

\section{}
Biped robots usually adopt feet with a rigid structure that simplifies walking on flat grounds and yet hinders ground adaptation in unstructured environments, thus jeopardizing stability. We recently explored in the SoftFoot the idea of adapting a robotic foot to ground irregularities along the sagittal plane.
Building on the previous results, we propose in this paper a novel robotic foot able to adapt both in the sagittal and frontal planes, similarly to the human foot. It features five parallel modules with intrinsic longitudinal adaptability that can be combined in many possible designs through optional rigid or elastic connections. By following a methodological design approach, we narrow down the design space to five candidate foot designs and implement them on a modular system. Prototypes are tested experimentally via controlled application of force, through a robotic arm, onto a sensorized plate endowed with different obstacles. Their performance is compared, using also a rigid foot and the previous SoftFoot as a baseline. Analysis of footprint stability shows that the introduction of the transverse arch, by elastically connecting the five parallel modules, is advantageous for obstacle negotiation, especially when obstacles are located under the forefoot. In addition to biped robots' locomotion, this finding might also benefit lower-limb prostheses design.
\tiny
\keyFont{\section{Keywords:} Robotic foot, adaptive foot, biped robots, humanoid robots, soft robotics, legged robots, biped locomotion}
\end{abstract}

\section{Introduction}

\begin{figure*}
 	\centering
 	\subfloat[][]
	{\includegraphics[width=0.20\textwidth]{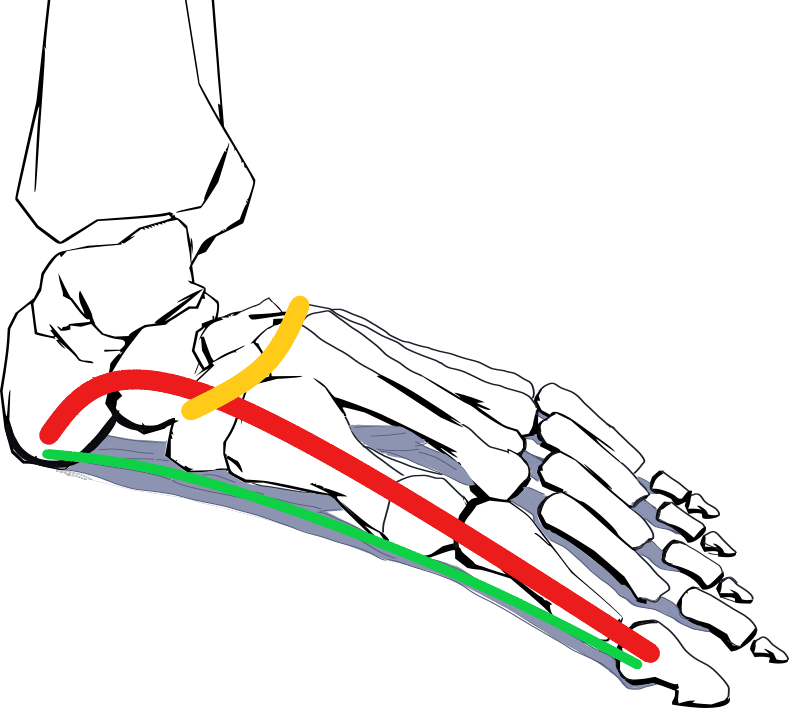}}\label{humanFoot} \quad
	\subfloat[][]
 	{\includegraphics[width=0.24\textwidth]{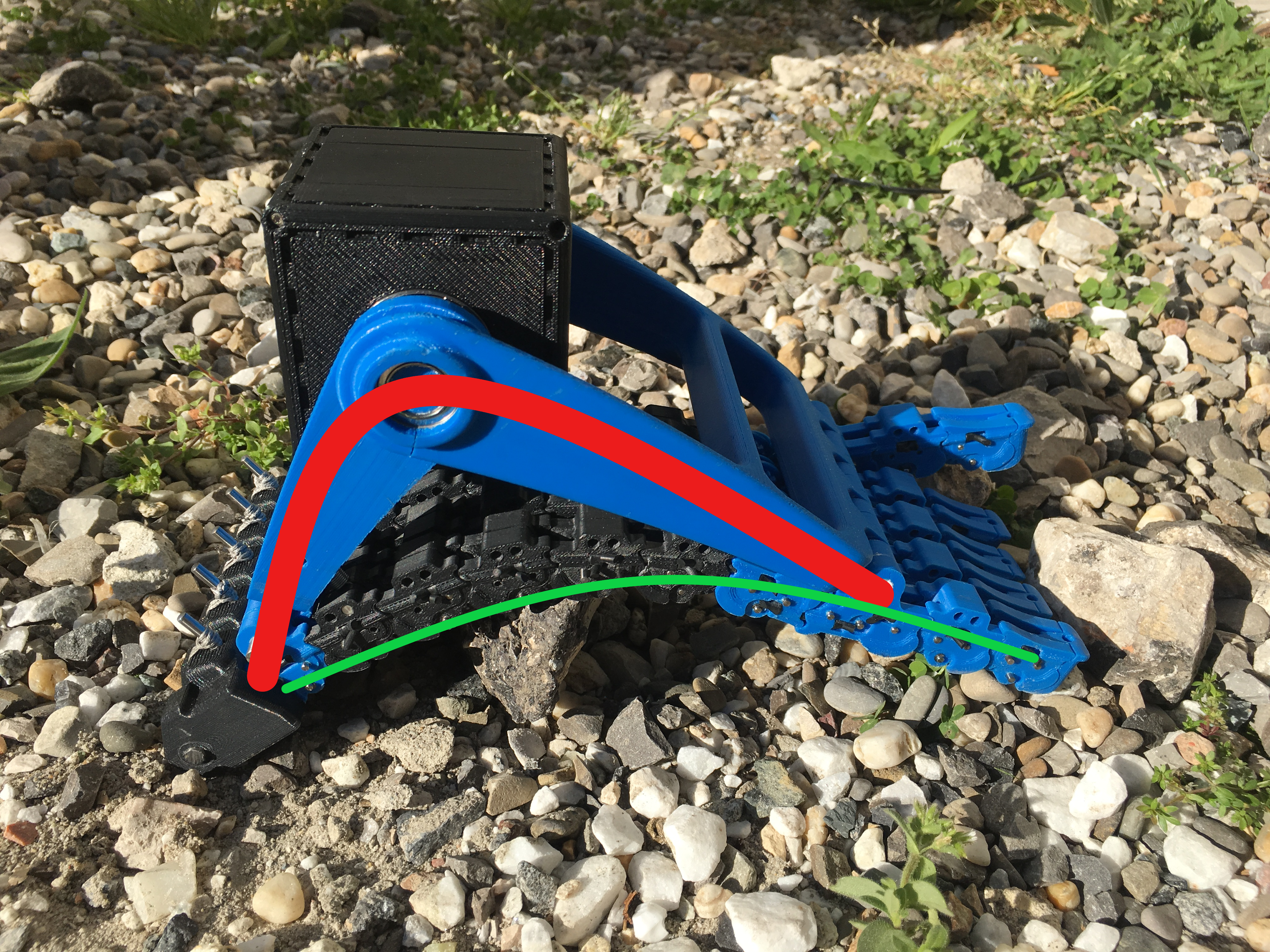}} \label{oldSoftFoot} \quad
    \subfloat[][]
 	{\includegraphics[width=0.29\textwidth]{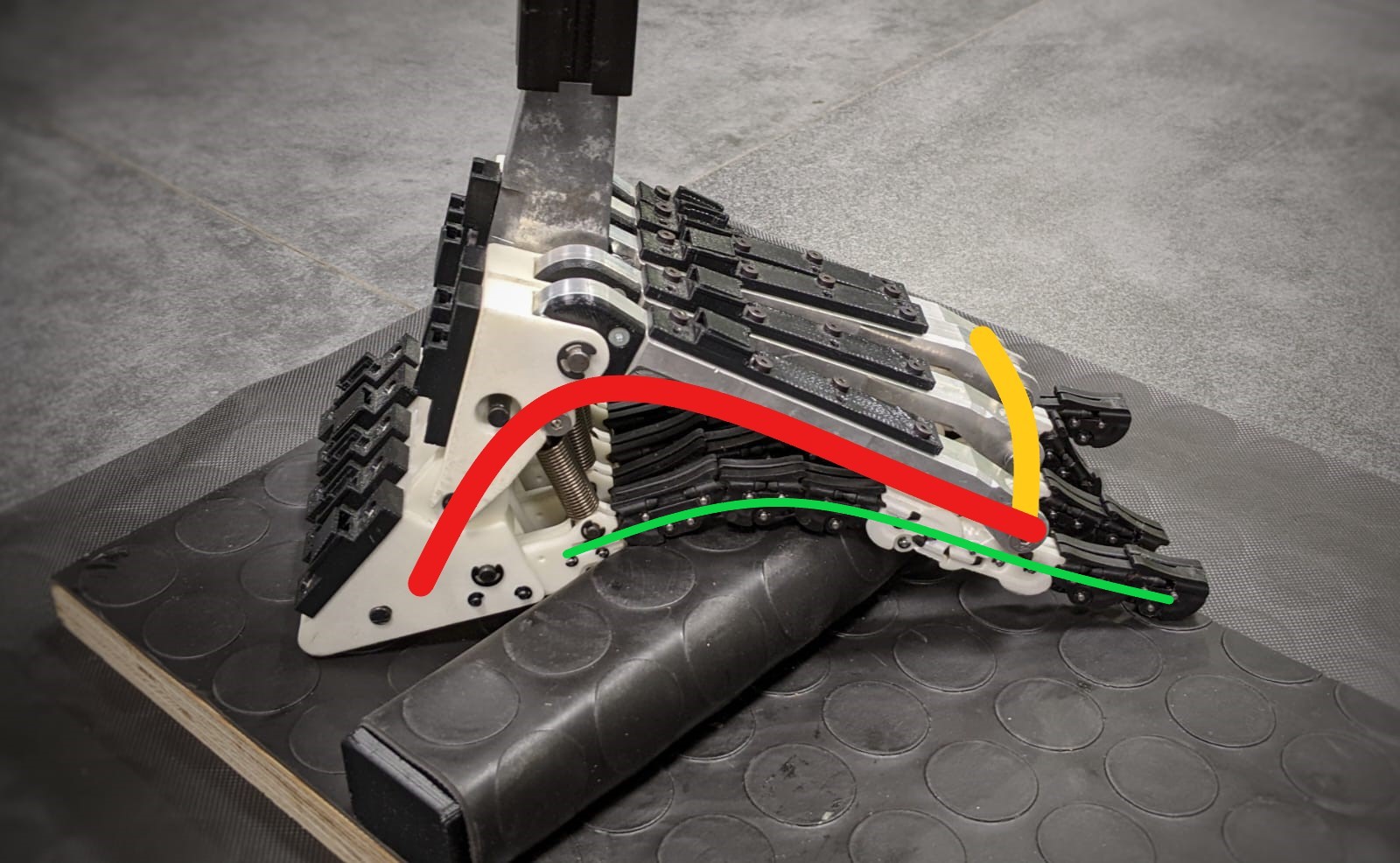}} \label{roboticFoot} 
 	\caption{\textbf{(A)} The human foot. \textbf{(B)} The SoftFoot described by \cite{piazza2016,piazza2024}. \textbf{(C)} The SoftFoot proposed in this paper. The longitudinal arch is highlighted in red, the transverse arch in yellow, and the plantar aponeurosis in green.}
    \label{humanRoboticFoot}
\end{figure*}

Locomotion represents a fundamental task for biped robots. Developed to operate in real-world scenarios, those robots should be able to navigate unstructured environments and handle disturbances. However, locomotion without falling on uneven surfaces is still a challenging task \citep{catalano2021}, and there is also still a lack of benchmarking of legged robot performance on irregular terrain \citep{torres2022}.

According to \cite{frizza2022}, most of the humanoid robots adopt, indeed, flat feet made of a single rigid element, which hinder ground adaptation, increasing the risk of falling. 
Most of the time falling prevention is addressed at the control level by developing algorithms to keep the humanoid robot balanced on diverse surfaces \citep{catalano2021, torres2022}.
Our research group, a few years ago, tackled this challenge from a different viewpoint, i.e. focusing on foot design, given its pivotal role in robot-ground interaction. We applied the embodied intelligence approach of soft robotics to design an adaptive soft robotic foot, i.e. the SoftFoot \citep{piazza2016, piazza2024} (see Figure \ref{humanRoboticFoot}B). It features an anthropomorphic and completely passive design exhibiting an intrinsic longitudinal ground adaptability. Our approach involved addressing the complexity of locomotion on irregular grounds at the mechanical level within the foot design, resulting in a simplification of control requirements.

The SoftFoot consists of two rigid components, a frontal and a rear arch, hinged at the ankle joint, and a sole with toes made of five parallel, flexible, and elastic structures connected through revolute joints to the rigid components (see \cite{piazza2016, piazza2024} for a detailed description of the foot architecture and working principle). The sole passively varies its shape and stiffness in the sagittal plane in response to the applied load and the ground profile, thanks to a peculiar mechanical structure combining rigid and elastic elements. 
When adopted by the humanoid robot HRP-4, the SoftFoot exhibited superior performance compared to the original flat feet of the robot \citep{catalano2021}, confirming the promising results obtained during preliminary bench tests on uneven ground \citep{piazza2016, piazza2024}.

Our soft robotic foot draws inspiration from the human counterpart, which modulates its stiffness within each gait cycle based on the environment and the task, guaranteeing impact absorption, adaptability, stability, and energy-efficiency \citep{venkadesan2017}. In the human foot, flexibility and softness at heel contact are combined with stiffness to convert the push-off ankle mechanical power into kinetic energy for forward body propulsion \citep{holowka2018,venkadesan2017,asghar2021}. The longitudinal arch, shown in red in Figure \ref{humanRoboticFoot}A, is the structure primarily responsible for the foot energy-recycling behaviour \citep{ker1987,holowka2018}, and it is usually associated to foot stiffness modulation \citep{holowka2018, venkadesan2020}. It consists of the tarsal and metatarsal bones kept together through ligaments, muscles, and the connective tissue of the foot base named \textit{plantar aponeurosis}. The longitudinal arch slightly flattens when loaded, absorbing impacts - together with the fat pad of the heel \citep{miller2002}, and adapting to unevenness \citep{kirby2017}. Meanwhile, the stretching of the plantar aponeurosis and ligaments allows energy storage, contributing to forward propulsion during foot unloading \citep{ker1987, mcdonald2016}, and to the arch stiffness modulation. Plantar aponeurosis and ligaments also ensure the foot structural integrity \citep{ker1987}, preventing an excessive flattening, while guaranteeing weight bearing and stability \citep{kirby2017}. In addition, push-off toes dorsiflexion engages a windlass-like mechanism \citep{venkadesan2020}: by further stretching the plantar aponeurosis, the calcaneus moves forward, causing the arch to rise, and the midfoot joints stiffen so that the foot acts as a lever, converting the power generated at the ankle into effective forward propulsion \citep{holowka2018}.

The two rigid components of the SoftFoot replicate the longitudinal arch, while the foot sole acts as the plantar aponeurosis, exhibiting an energy-recycling mechanism through the stretching of the elastic components when loaded, compliantly conforming to the ground, and absorbing impact while keeping the integrity of the foot structure. In the sole, high compliance (i.e. a large number of degrees of freedom (DoFs)) for ground adaptability and stiffness required for stability coexist. This was proved in \cite{piazza2024} by analytically comparing the SoftFoot with both a rigid and a completely compliant foot (e.g. a foot with a layer of soft material under the sole) in terms of stability and adaptation.

In research, some adaptive designs exist that incorporate compliant elements like rubber at the bottom of the rigid sole for impact absorption, while others enhance obstacle negotiation by adding DoFs through rigid segments connected through revolute joints \citep{frizza2022}. Examples of multi-DoFs and multi-joints structures are those described in \cite{davis2010, kuehn2012, asano2016}, as well as the SoftFoot. Another attempt is the active robotic foot by \cite{qaiser2017}, which is supposed to adapt in real-time to the ground by actively tuning the stiffness of the two concentric helical springs (by changing the number of active coils) working as the plantar aponeurosis.

However, those designs - including the SoftFoot - primarily exhibit adaptation in the sagittal plane. Attempts to enhance adaptability beyond the sagittal plane include the design by \cite{kawakami2015}, revised in \cite{enomoto2022}, with an oblique axis working as the midfoot joint and allowing some forefoot abduction/adduction, and its further evolution providing motion in the frontal plane \citep{chen2022}. In addition, the tethered robotic ankle-foot prosthesis emulator described in \cite{collins2015} and in \cite{chiu2021}, with separate control of the three digits - two for the forefoot and one for the hindfoot, allows frontal and sagittal plane adaptation. Just like \cite{seo2009}'s robotic foot, with five rigid toes hinged at five rigid metatarsal segments, a wheeled rigid heel, and a sole made of springs aiming to replicate the plantar aponeurosis.

The SoftFoot stands out for the highest number of DoFs and for being completely passive \citep{frizza2022}. Nevertheless, the five structures of its sole can move only partially with respect to each other in the frontal plane, because of the transverse axis connecting them to the frontal arch. To address this limitation, the work presented in this paper investigates the incorporation of a flexible transverse arch into the SoftFoot design (Figure \ref{humanRoboticFoot}C). Inspired also by a recent study on the human foot by \cite{venkadesan2020}, which shows how the transverse tarsal arch (Figure \ref{humanRoboticFoot}A) affects sagittal foot stiffness thanks to the stiffening of the inter-metatarsal tissues, we explore the changes in the SoftFoot performance when a flexible transverse arch is introduced. We hypothesize that, in this way, the adaptability of the SoftFoot can be extended to ground variations in the frontal plane, ultimately enhancing stability on unstructured terrains.

This paper details the replication of the transverse arch function in the SoftFoot and narrows down the design space to five classes in Section 2. The performance evaluation, outlined in Section 2, includes benchtop testing with a robotic arm applying forces to the prototypes in contact with a plate with obstacles of different sizes. The performance of the SoftFoot with both arches on unevenness is compared to that of the previous SoftFoot and of a rigid foot in terms of stability, being this the primary goal of increasing foot adaptability to the ground. Results are presented in Section 3 and discussed in Section 4. 
\section{Materials and methods}
\subsection{Mechanical design of the SoftFoot 3D}

\newlist{mycons}{enumerate}{2}
\setlist[mycons, 1]
{nosep,label=C\arabic{myconsi}., 
leftmargin=30pt}

\setlist[mycons, 2]
{label=C\arabic{myconsi}.\arabic{myconsii}, 
leftmargin=24pt}

\newlist{myassumptions}{enumerate}{1}
\setlist[myassumptions, 1]
{label=A\arabic{myassumptionsi}.,leftmargin=30pt}

\begin{figure*}[!h]
 	\centering
 	\includegraphics[width=0.8\textwidth]{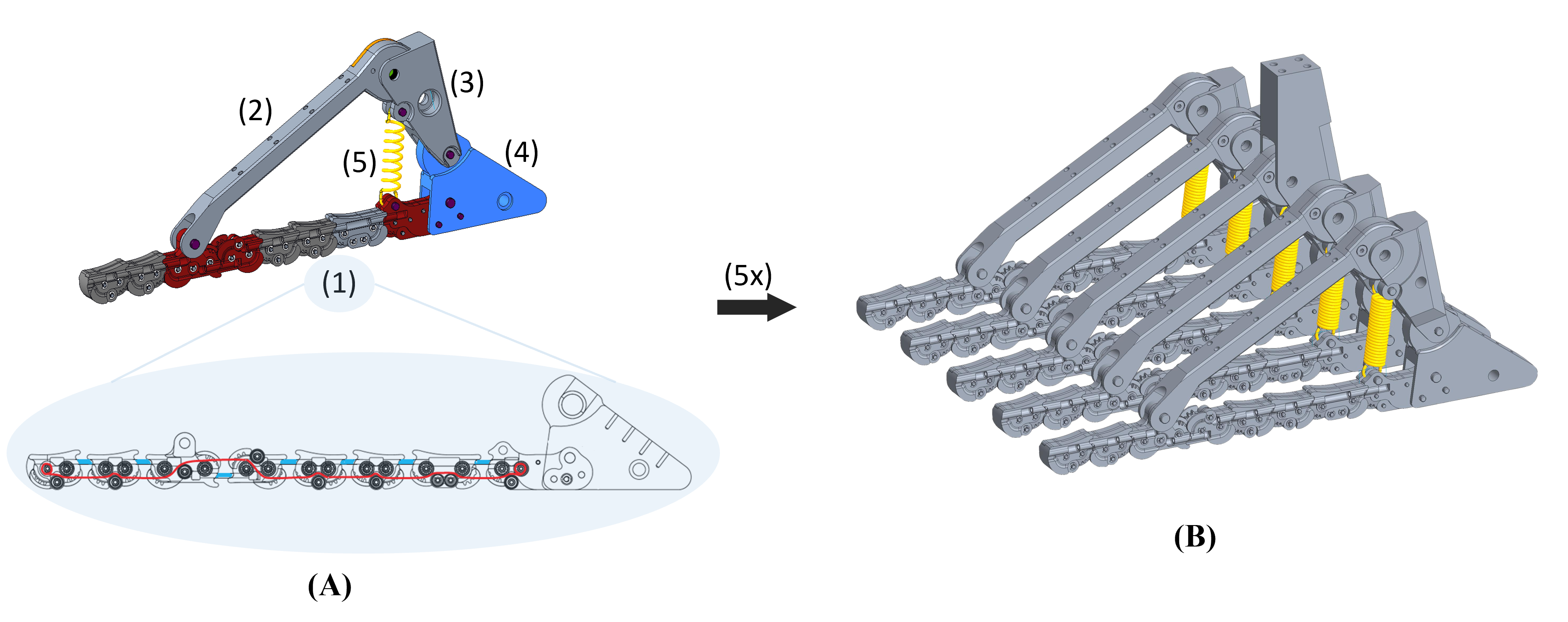}
	\caption{\textbf{(A)} The components  of the basic 2D module are: (1) adaptive sole; (2) frontal arch; (3) rear arch; (4) heel; (5) coil spring. In the adaptive sole, the elastic bands are shown in light blue, and the tendon routing in red. \textbf{(B)} Five basic 2D modules placed in parallel form the SoftFoot 3D. Note that the rear arch of the central module is longer than that of the other modules to allow the connection of the foot to the user and, thus, forces transmission.}
    \label{basicto3D}
\end{figure*}

The structure of the SoftFoot 3D\footnote{The SoftFoot under evaluation in this study is sometimes referred to as \textit{SoftFoot 3D} within this paper to highlight the transition from adaptability in the sagittal plane only (exhibited by each \textit{2D module}) to adaptability on both the frontal and sagittal planes.} consists of five basic 2D modules placed in parallel, each one made of four main components: an adaptive sole, a frontal arch, a rear arch, and a heel (Figure \ref{basicto3D}A). The adaptive sole is in turn made of eight small rigid bodies connected by one elastic and one inextensible link. A coil spring (spring constant 0.125 N/mm) connects the rear arch of each module to the rearmost small rigid body of the adaptive sole, which in turn is rigidly connected to the heel component (Figure \ref{basicto3D}A). All rigid bodies of each 2D module are 3D printed.

\begin{figure*}[!ht]
 	\centering
 	\includegraphics[width=1\textwidth]{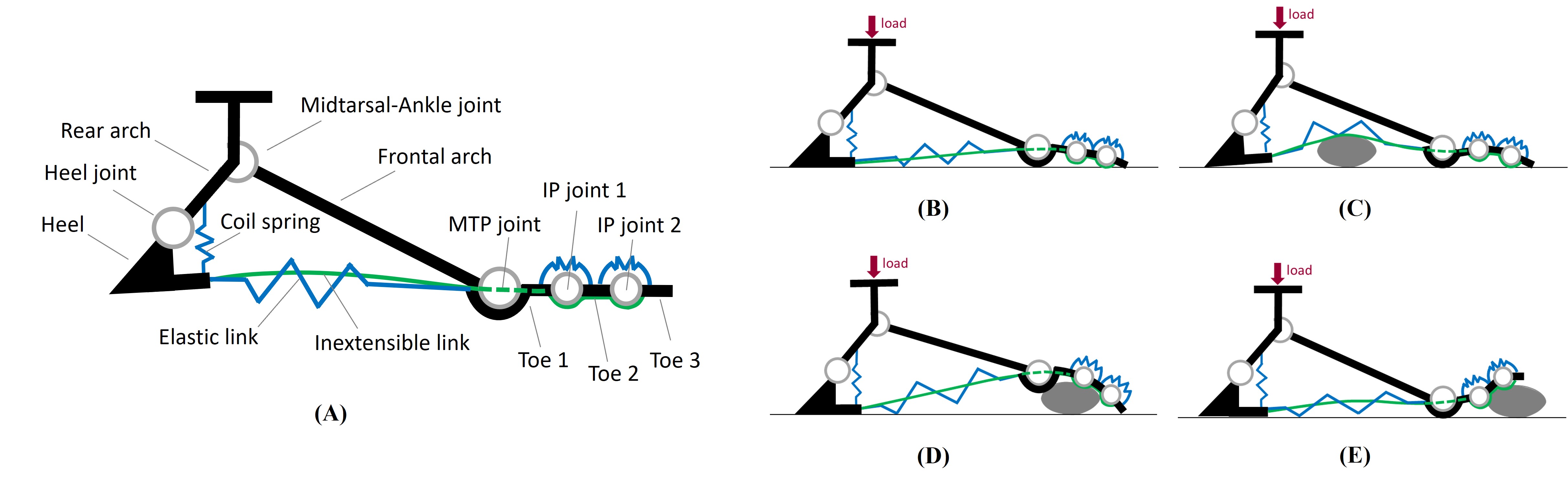}
\caption{\textbf{(A)} Schematic representation of the six components (referred to as \textit{links}) and the five joints of each basic module of the SofFoot 3D. The coil spring connecting the rearmost small rigid body of the adaptive sole to the rear arch is also shown, together with the two elastic elements at the toes representing the passive elastic joints connecting the three phalanges in series. The smaller images on the right show the behaviour of the adaptive sole when the foot is loaded \textbf{(B)} on even ground, or \textbf{(C)-(E)} on obstacles.} 
    \label{dofss}
\end{figure*}

\subsubsection{The 2D module}
A schematic representation of the basic 2D module of the SoftFoot 3D is given in Figure \ref{dofss}A. The components of each module (also referred to as \textit{links}) are shown in black: three phalanges (\textit{Toe 1}, \textit{Toe 2}, and \textit{Toe 3}), the \textit{frontal arch}, the \textit{rear arch}, and the \textit{heel}. The joints are also shown: two interphalangeal (IP) joints between the toes (i.e. \textit{IP joint 1} and \textit{IP joint 2}), the \textit{MTP joint}, the \textit{midtarsal-ankle joint}, and the \textit{heel joint}. Note that, although the names of the links and the joints evoke an anatomical parallelism, this has purely a mnemonic purpose \footnote{For instance, Toe 1 is connected to the frontal arch with a pin joint, the midtarsal-ankle joint represents a hybrid between these two joints in the human foot.}. Indeed, our design does not try to merely mimic human anatomy, but rather to reproduce the features of the human foot function.

The \textit{Elastic link} consists of a pair of elastic bands (natural rubber, diameter 2.5 mm, length 10 mm, nominal stiffness 35 ShA) connecting each small rigid body of the sole to the following one (Figure \ref{basicto3D}A). They are the same already used to connect the phalanges of the SoftHand v1.2 for research applications\footnote{\url{https://qbrobotics.com/product/qb-softhand-research/}}. They allow the foot sole to go back to its original shape after deformation. The \textit{Inextensible link} is realized through a tendon (Liros DC000-0200, i.e. pre-stretched coated Dyneema SK 75) connecting all the small bodies of the sole, from the rearmost one to the toe tip, according to a specific routing within them for a proper load distribution (Figure \ref{basicto3D}A). The tendon was pre-tensioned considering the maximum tension acting on the foot sole during push-off in a 100 kg user - deriving from the corresponding maximum ankle joint torque \citep{cherelle2014}, an equal force distribution among the five 2D modules, and a safety coefficient equal to 2. This choice complies also with the maximum tension measured in the plantar aponeurosis across its five slips during push-off \citep{caravaggi2009}: about 1.5 times body weight, resulting in almost 300 N in each slip in a 100 kg person - if an equal load distribution is applied. The elastic and inextensible links work together in parallel to provide a stiffening-by-compression behavior: the sole goes from a compliant to a stiff behavior when the load on the foot increases, withstanding the user's load while ensuring ground adaptability (see \cite{piazza2024} for a complete explanation). Specifically, the elastic bands and the tendon stretch under the user's load, tightening the adaptive sole. This mechanism replicates what happens in the human foot during the stance phase of gait (i.e. when the foot is in contact with the ground), when the connective tissue of the \textit{plantar aponeurosis}, connecting the calcaneus to the metatarsal heads, elongates under the body weight, providing stiffness. During such a stretching action, thanks to an appropriate placement of the two aforementioned links across the eight rigid bodies, the sole of the SoftFoot tends to assume a curved shape, which is more evident at the forefoot level, with the toes exercising a pushing force on the surface (Figure \ref{dofss}B). This peculiar design of the adaptive sole is what allows filtering ground unevenness (Figure \ref{dofss}C-E). In addition, during push-off, Toe 2 and Toe 3 dorsiflexion leads the adaptive sole to stretch, replicating the windlass mechanism. In this phase of the gait cycle, the pushing force of the toes on the ground provides a stable support area for forward body propulsion. 

The stiffness of the coil spring is determined to prevent an excessive backward rotation of the rear arch with respect to the heel when the foot is loaded in the hindfoot area, which would cause user's instability, and to limit the closure motion between rear arch and sole. Moreover, the spring works as a shock absorber reducing the impact forces transmitted to the user. It keeps the adaptive sole slightly curved in the unloaded condition, whereas both the spring and the sole are stretched when loaded.

\subsubsection{From the 2D module to the SoftFoot 3D} 
We systematically investigate the connection between five parallel basic modules  with the goal of achieving the best adaptability also in the frontal plane. We consider three possible types of connections across the 6 links and the 5 joints: \textit{free} (F), \textit{elastic} (K), and \textit{rigid} (R). We also introduce some preliminary \emph{assumptions} to simplify the problem complexity:
\begin{myassumptions}
    \item\ Forces and torques are transmitted from the foot to the user's leg through the rear arch of the central module, via a rigid connection.
    \item\ For implementation simplicity, each link or joint can be connected only to its analog in the adjacent modules.
    \item\ To reduce the design space, the five modules are connected to each other in the same way.
    \item\ The modules move relatively to one other only in the sagittal plane.
\end{myassumptions}

Each link of a module is free to rotate and translate in-plane with respect to its analog in the adjacent module if no transversal connection exists among the five modules. This means that each link is characterized by three DoFs (one rotation and two translations) that can be constrained according to the three aforementioned connection types. Each joint, being of the revolute type, presents instead two constrainable DoFs corresponding to the two translations. 

Next, we make some sensible functional, kinematic and implementational \emph{considerations} to decrease the number of combinations that can be generated by connecting all the links and joints according to the three connection types:
\begin{mycons}
    \item\ The connection type of the two constrainable translational DoFs for each link and for each joint must be the same, to avoid increasing foot size and complexity by practically implementing different connections in the two directions.
    \item\ The IP joint 1 and IP joint 2 have an independent kinematics with respect to the other joints and, thus, are not considered.
    \item\ The connections among the three toes are set as \textit{free} for their three DoFs, because their small dimension makes difficult to connect them in other ways.
    \item\ The frontal arches can be only elastically connected (valid for both translations and rotation): indeed, a \textit{rigid} connection would prevent frontal adaptability, while a \textit{free} connection might lead to excessive displacements and, thus, structural instability, which is risky for the user, especially during push-off.
    \item\ The rear arches cannot be \textit{free}, since they transmit motion between the user and the foot.
    \item\ The heel must have the same connection type for translations and rotation, to avoid bulky and complex connections when different types are considered.
    \item\ The midtarsal-ankle joint is \textit{rigid} to transmit most of the user's load. \par This point has the following consequences:
    \begin{mycons}
            \item\ The rear arch and the frontal arch can only pivot on it, so they retain only one rotational DoF.
            \item\ An \textit{elastic} connection on the frontal arch is equivalent to one on the MTP joint, so we consider the frontal arches as elastically connected, and the MTP joints as \textit{free}, for implementation simplicity.
            \item\ Likewise, an \textit{elastic} connection on the rear arches is equivalent to one on the heel joints. Therefore, we connect elastically the rear arches, and let the heel joints \textit{free}.
            \item\ Similarly, a \textit{rigid} connection on the rear arches is equivalent to one on the heel joints, given that they will anyway behave as a unique rigid body. Therefore, we consider the rear arches as being rigidly connected, while the heel joints as \textit{free}.
            \item\ If the rear arches are rigidly connected, the heel can only rotate around the heel joint, so that it has only one DoF with respect to the adjacent heels.
            \item\ If the heels are rigidly connected, the connection between rear arches and heel joints is indifferent. The rear arch, indeed, can rotate only rigidly around the heel joint with respect to the heel, with a \textit{free} heel joint given considerations C7.3 and C7.4.
    \end{mycons}
\end{mycons}

\begin{figure}[hb!]
	\centering
	\subfloat[][\small\textbf{KKF}]
    {\includegraphics[width=0.20\columnwidth]{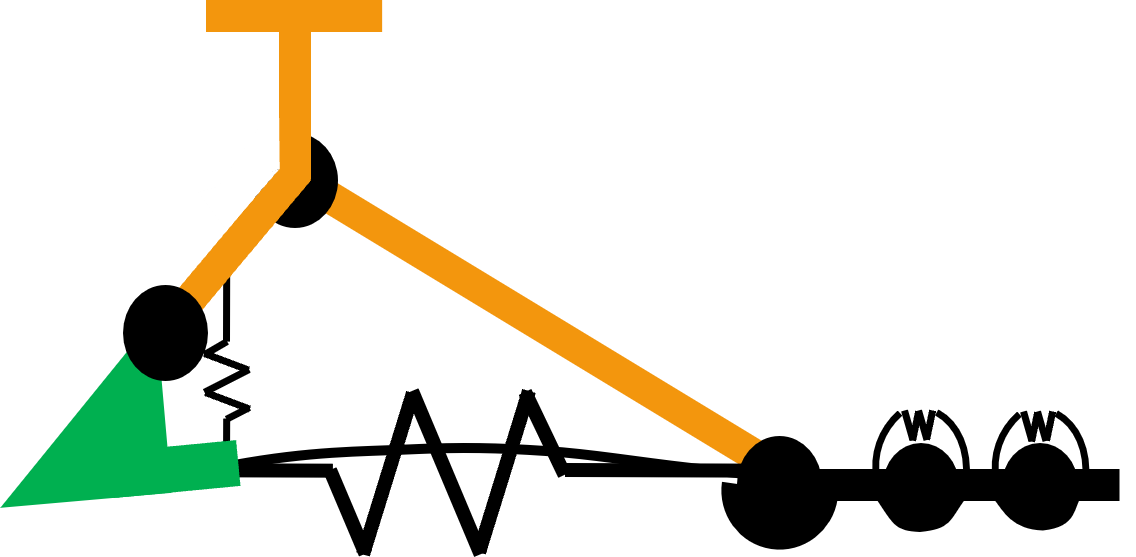} \label{}} \quad
	\subfloat[][\small\textbf{KKK}]
	{\includegraphics[width=0.20\columnwidth]{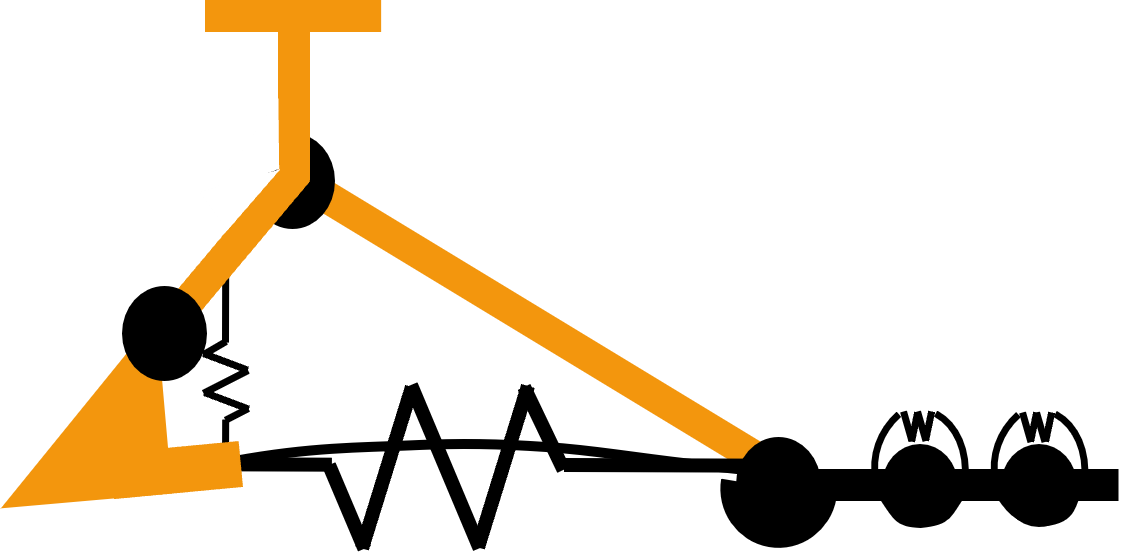} \label{}} \quad
	\subfloat[][\small\textbf{KRF}]
	{\includegraphics[width=0.20\columnwidth]{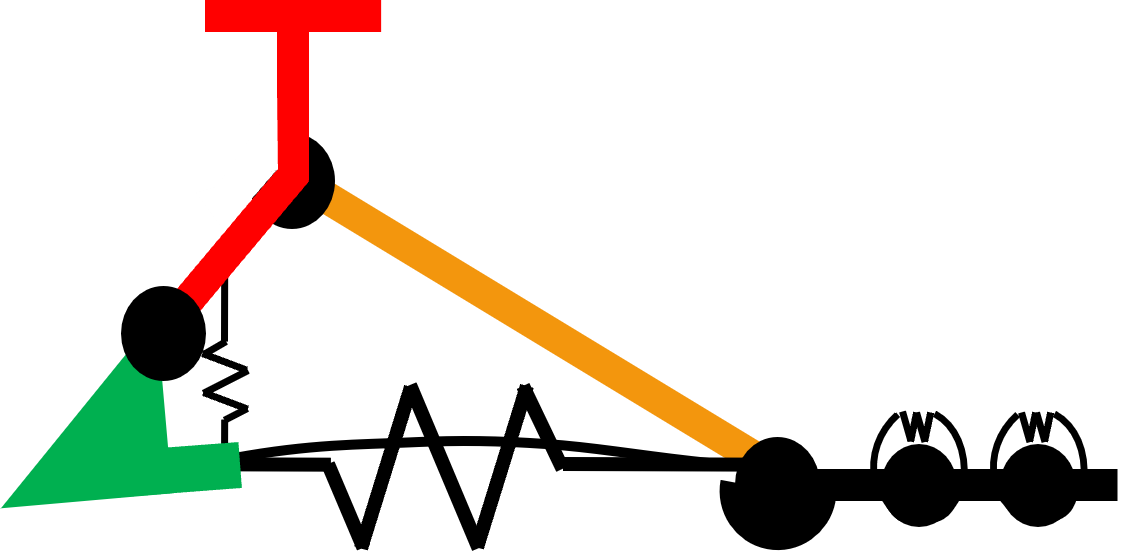} \label{}} \\
	\subfloat[][\small\textbf{KRK}]
	{\includegraphics[width=0.20\columnwidth]{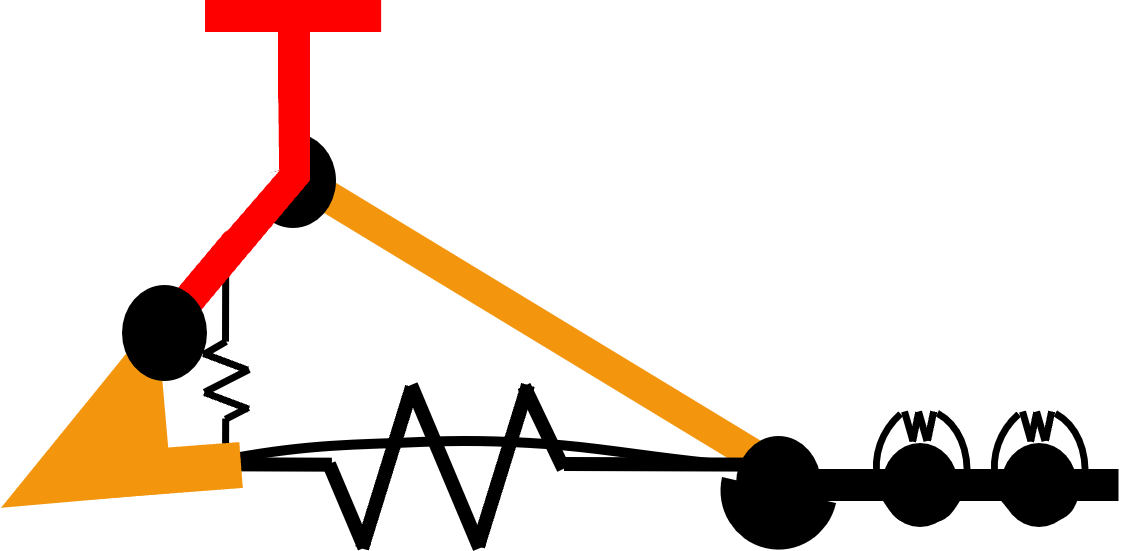} \label{}} \quad
	\subfloat[][\small\textbf{KRR}]
	{\includegraphics[width=0.20\columnwidth]{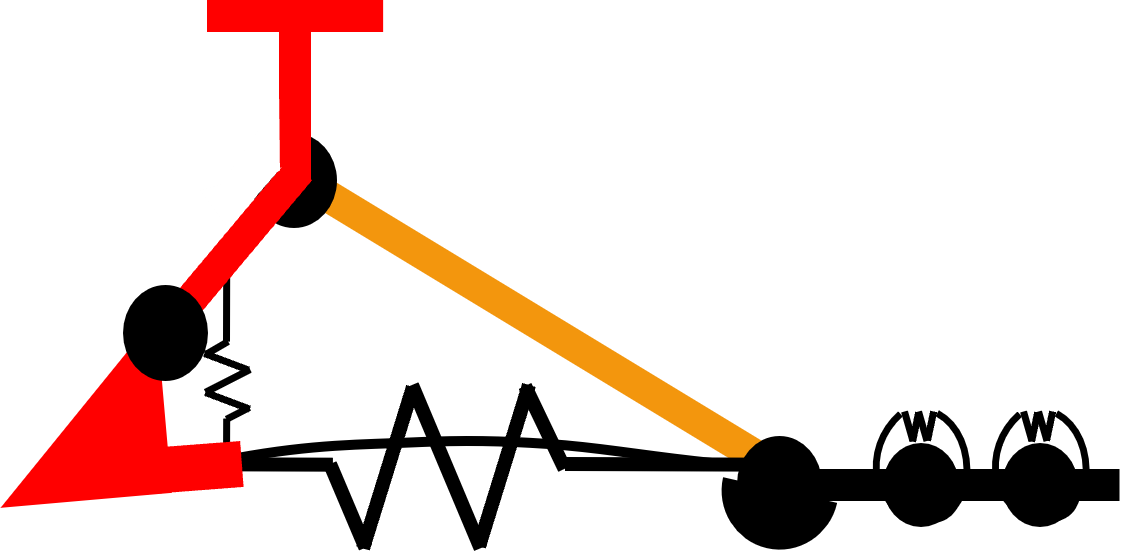} \label{}} \quad
	\subfloat[][\small\textbf{RRR}]
	{\includegraphics[width=0.20\columnwidth]{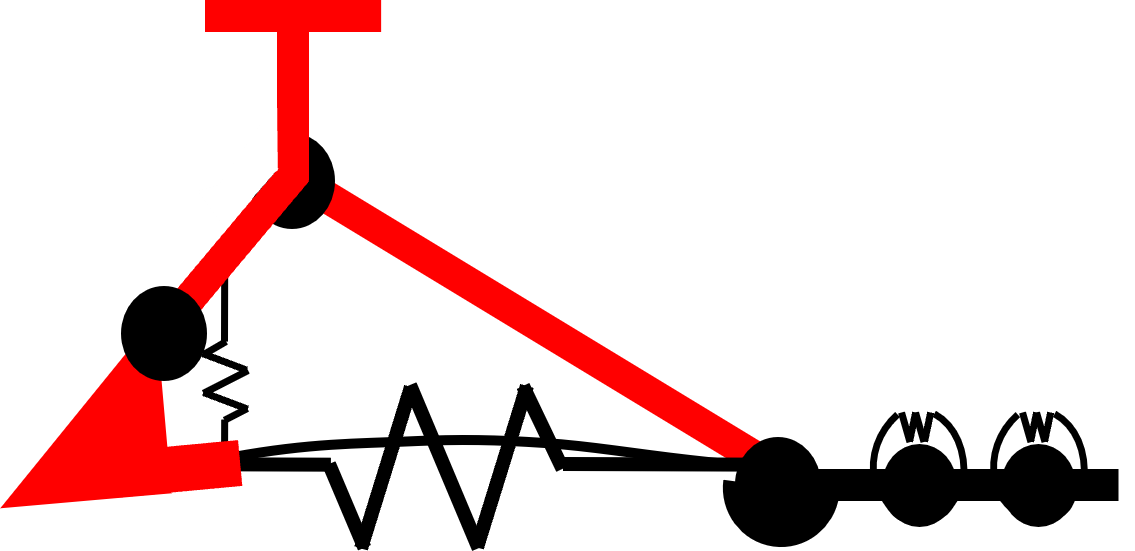} \label{}} \\
	\subfloat
	\centering
	{\includegraphics[width=0.5\columnwidth]{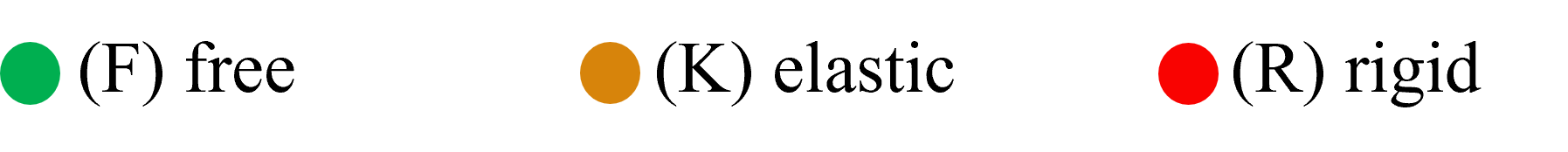} \label{}} \\
    \caption{\textbf{(A)-(E)} The five selected configurations of the SoftFoot 3D and \textbf{(F)} the SoftFoot described in \cite{piazza2016, piazza2024}, labeled according to the connection types across respectively the frontal arches, the rear arches, and the heel links. The connection types are displayed in green if \textit{free} (F), orange if \textit{elastic} (K), and red if \textit{rigid} (R).}
    \label{conf}
\end{figure}

As a result of all the previous assumptions and considerations, the possible connections left across links and joints are those displayed in Table \ref{possibilities}. The only components having different connection options are the rear arches and the heels. Specifically, both of them can be elastically or rigidly connected, and the heels can also be left \textit{free}, resulting in 6 possible combinations. We label the possible solutions by using three letters corresponding to the connection type for the DoFs of respectively the frontal arches, the rear arches, and the heels. The six combinations are: A) KKF, B) KKK, C) KKR, D) KRF, E) KRK, F) KRR. Finally, since KKR is equivalent to combination KRR under consideration C7.6, it can be excluded. Therefore, five final combinations can be realized by connecting the rear arches and the heels in various ways: A) KKF, B) KKK, C) KRF, D) KRK, E) KRR (Figure \ref{conf}A-E).
\begin{table}[ht!]
 \caption{Feasible connection types for the DoFs of links and joints of each module, as a result of the assumptions and considerations made (the latter are displayed in the third column).\vspace{3mm}} 
    \label{possibilities}
    \centering
    \begin{tabular}{|c|c|c|}
    \hline
     \textbf{Components} & \textbf{Connections} &
     \textbf{Considerations} \\
     \hline
     Toe3, Toe2, Toe1 & F & C3\\
     \hline
     MTP joint & F & C7.2\\
     \hline
     Frontal arch & K & C4\\ 
     \hline
     Midtarsal-Ankle joint & R & C7\\ 
     \hline
     Rear arch & K, R & C5\\
     \hline
     Heel joint & F & C7.3, C7.4 \\
     \hline
     Heel & F, K, R & C6\\
     \hline
    \end{tabular}
\end{table}

\begin{figure}[h!]
	\centering
	\subfloat[][\small\textbf{KKF}]
 	{\includegraphics[width=0.20\textwidth]{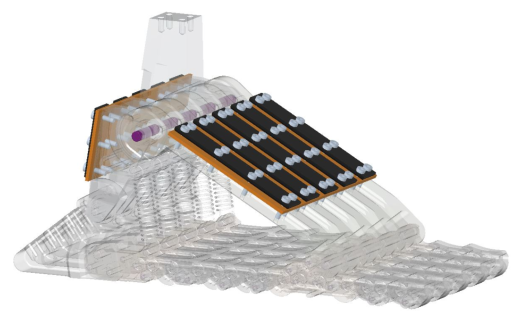} \label{}} \quad
	\subfloat[][\small\textbf{KKK}]
	{\includegraphics[width=0.20\textwidth]{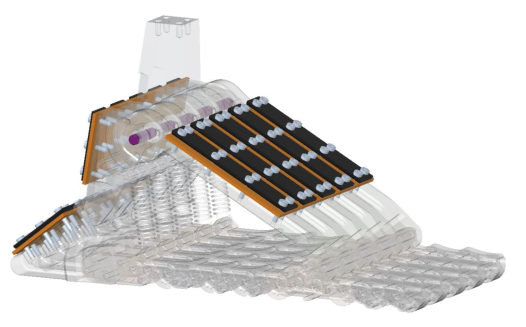} \label{}} \quad
	\subfloat[][\small\textbf{KRF}]
	{\includegraphics[width=0.20\textwidth]{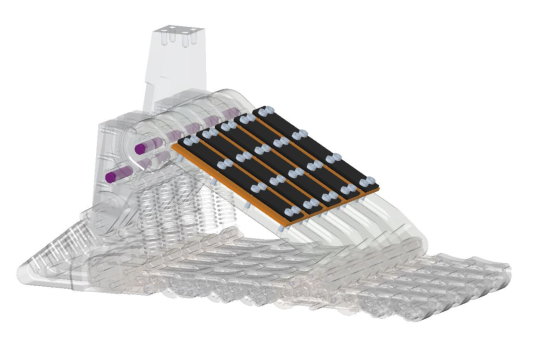} \label{}} \quad 
	\subfloat[][\small\textbf{KRK}]
	{\includegraphics[width=0.20\textwidth]{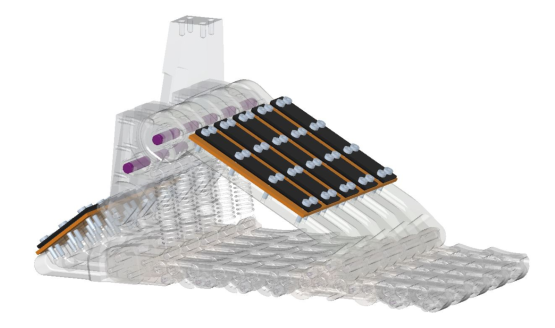} \label{}} \quad
	\subfloat[][\small\textbf{KRR}]
	{\includegraphics[width=0.20\textwidth]{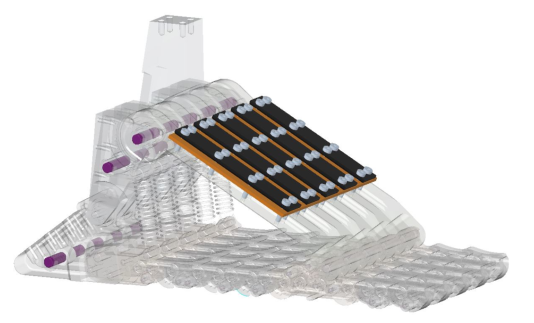} \label{}} \quad
	\subfloat[][\small\textbf{RRR}]
	{\includegraphics[width=0.20\textwidth]{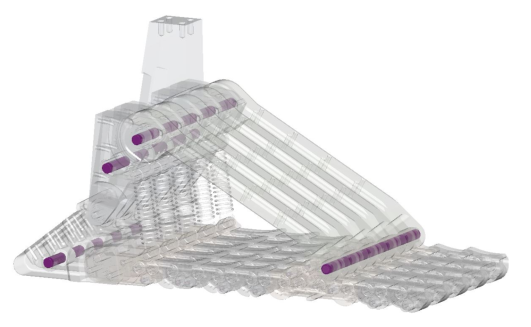} \label{}} \quad
	\subfloat[][\small\textbf{Rigid foot}]
	{\includegraphics[width=0.24\textwidth]{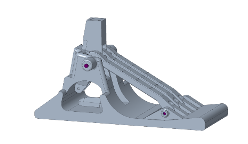} \label{}} \quad
	\caption{Solid models of \textbf{(A)-(E)} the five configurations of the SoftFoot 3D, \textbf{(F)} the SoftFoot described in \cite{piazza2016, piazza2024}, and \textbf{(G)} a completely rigid foot. The constraints to connect the five basic modules are implemented: cylindrical steel pins are used for the rigid connection; a sheet of nitric rubber fixed to the links through bolts is used for the elastic connection.}
    \label{dofs}
\end{figure}

The practical implementation of the \textit{rigid} and the \textit{elastic} connections is defined based on the following requirements: modularity, lightweight, small volume, simplicity in assembly and manufacturing, low-cost. As a result, the \textit{rigid} connections are implemented through cylindrical steel pins, and the \textit{elastic} ones using sheets of nitric rubber (shore A50, 2 mm thick) connected through bolts to the links. The thickness is heuristically selected as a compromise in between a larger one - resulting in a stiff behaviour - and a smaller one - resulting in a weak connection between 2D modules. Figure \ref{dofs}A-E shows the corresponding solid model of the five configurations.\\
\subsection{Experimental Testing}

\newlist{myitems}{enumerate}{1}
\setlist[myitems, 1]
{label=\arabic{myitemsi}.,leftmargin=23pt}

The stability performance of the five configurations of the SoftFoot 3D when loaded on unevenness is experimentally evaluated. The primary function of a foot, for both human beings and bipedal robots, is to provide a stable support during standing and walking, both on flat and uneven grounds. The stability of some legged systems on even grounds can be effectively defined by using the Zero-Moment Point (ZMP) criterion, but difficulties arise when dealing with uneven grounds \citep{sardain2004}. The definition of the ZMP concept and its application to the stability control of a bipedal system requires that both feet are in contact with the same ground surface having one associated normal vector.
When this condition is lacking, e.g. in the case of uneven terrains, leading the two feet to be in contact with different ground surfaces having different normal vectors, the ZMP criterion cannot be naively applied. For this reason, we experimentally evaluate the stability of these adaptive feet on uneven ground through benchtop testing.

\subsubsection{Setup} \label{sssec:num1}
\begin{figure*}[hb!]
\centering
{\includegraphics[width=0.7\textwidth]{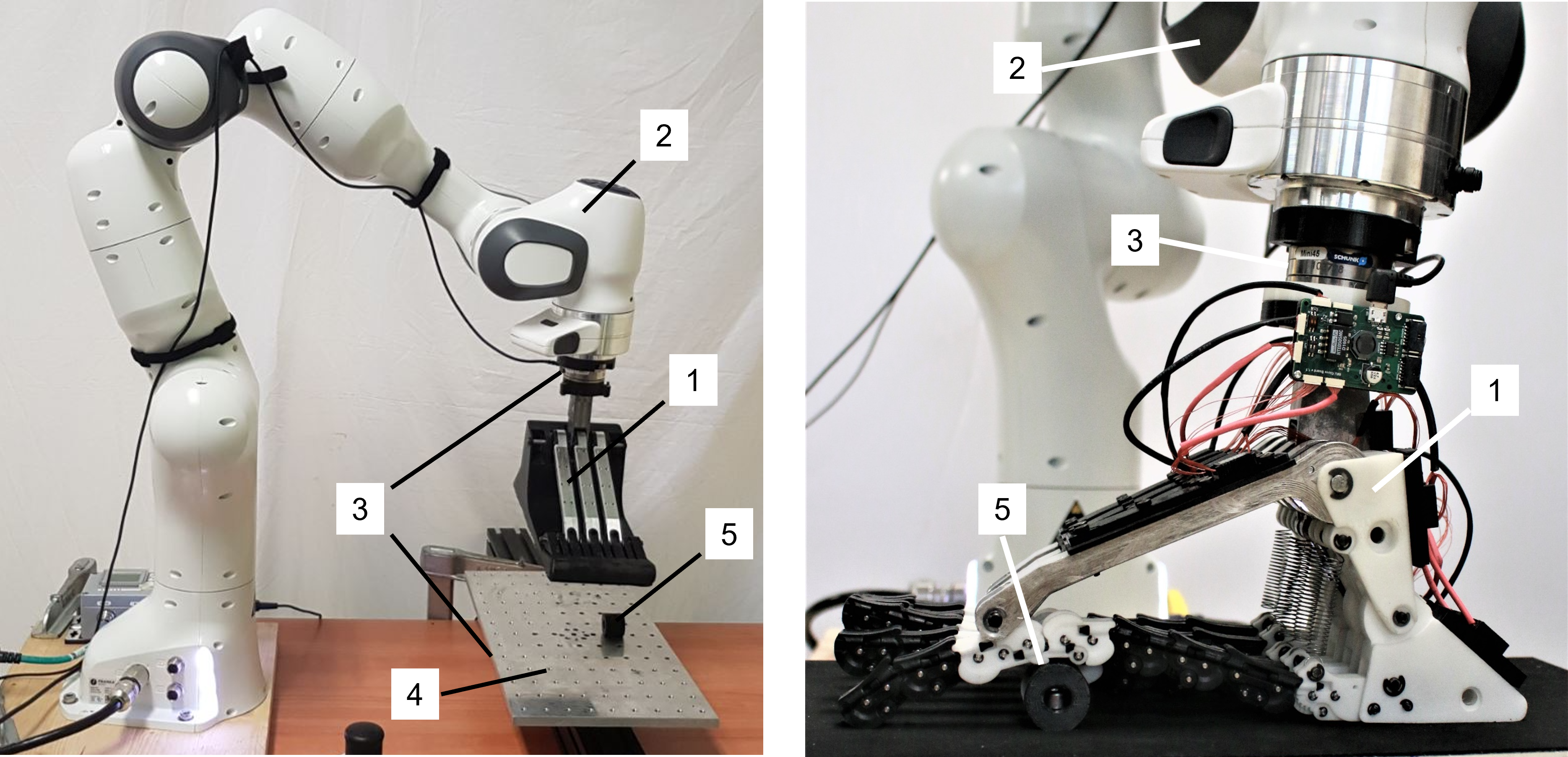}} 
\caption{Experimental setup: (1) the foot to be tested; (2) the robotic arm; (3) 6-axes force sensors; (4) aluminum plate with threaded holes; (5) 3D printed cylindrical obstacle. Fifteen Inertial Measurement Units (IMUs) are also visible in the image on the right. They are mounted on the frontal arch, rear arch, and heel of each 2D module of the SoftFoot. However, the data from the IMUs are not used in this study.}
\label{setup}
\end{figure*}

Two more foot configurations are considered for comparison:
\begin{enumerate}
    \item\ A configuration replicating the SoftFoot described in \cite{piazza2016, piazza2024}, featuring only longitudinal adaptability, and implemented through rigid connections at the MTP joints, rear arches, and heels. It is labeled as RRR configuration, given the equivalence of a rigid connection on the MTP joint and the frontal arch  (Figures \ref{conf}F and \ref{dofs}F for the solid model).
    \item\ A fully rigid foot made of a 3D-printed ABS plastic part (which serves as the heel, plantar aponeurosis, and forefoot), three frontal arches, and a central rear arch, all rigidly connected with steel pins (Figure \ref{dofs}G, and also showed on the left of Figure \ref{setup}). It represents the benchmark for robotic feet. It is designed to have its mass and center of mass almost coincident with those of the other configurations, to allow for a direct comparison of the results. 
\end{enumerate}

The experimental bench includes:
\begin{itemize}
	\item\ the foot prototype to be tested (Figure \ref{setup} (1));
	\item\ a 7 DoFs anthropomorphic robotic arm (Franka Emika Panda) with a payload of 3 kg and its controller (Figure \ref{setup} (2));
	\item\ two 6-axes force sensors (ATI Mini 45-E sensors) (Figure \ref{setup} (3));
	\item\ an aluminum plate (400x200 mm) with a pattern of threaded holes (13x7 holes for a total of 91 holes) (Figure \ref{setup} (4));
	\item\ 3D printed cylindrical obstacles (diameter of 25 mm and length of 22 mm) (Figure \ref{setup} (5)).
\end{itemize}

The foot is mounted at the end effector (EE) of the robotic arm, through which the force is applied to the prototype, with an ATI force sensor (named "upper ATI") placed in between to calculate the wrench at the EE. An aluminum plate simulates the ground, with a second ATI force sensor (named "lower ATI") placed right below it. The pattern of holes allows to screw obstacles to the plate to simulate ground unevenness. A 1 mm thick layer of neoprene rubber is used to cover both the plate and the obstacles to increase friction.

\begin{figure*}[h]
	\centering
	\subfloat[]
	{\includegraphics[width=0.5\textwidth]{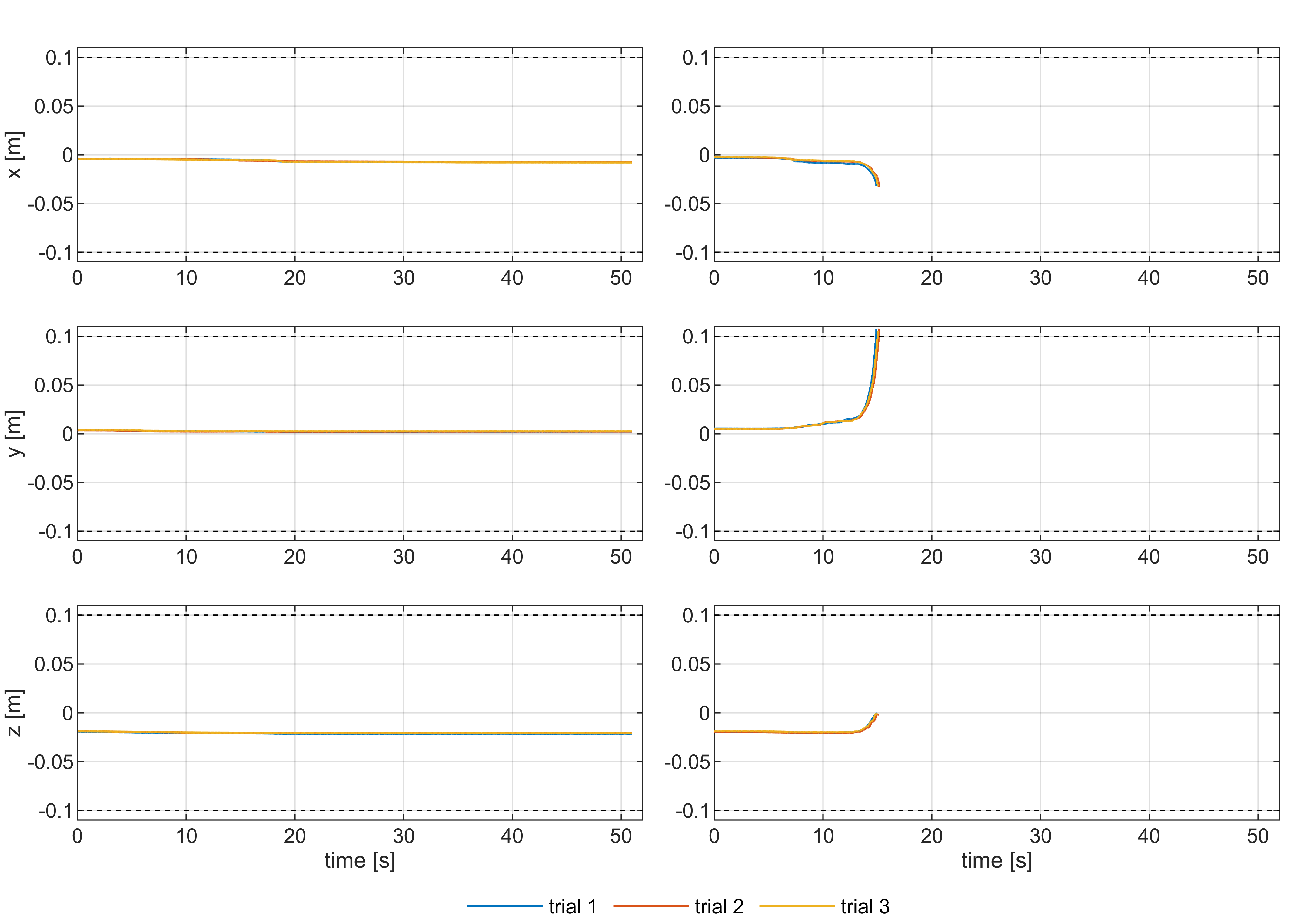}}
	\subfloat[]
	{\includegraphics[width=0.5\textwidth]{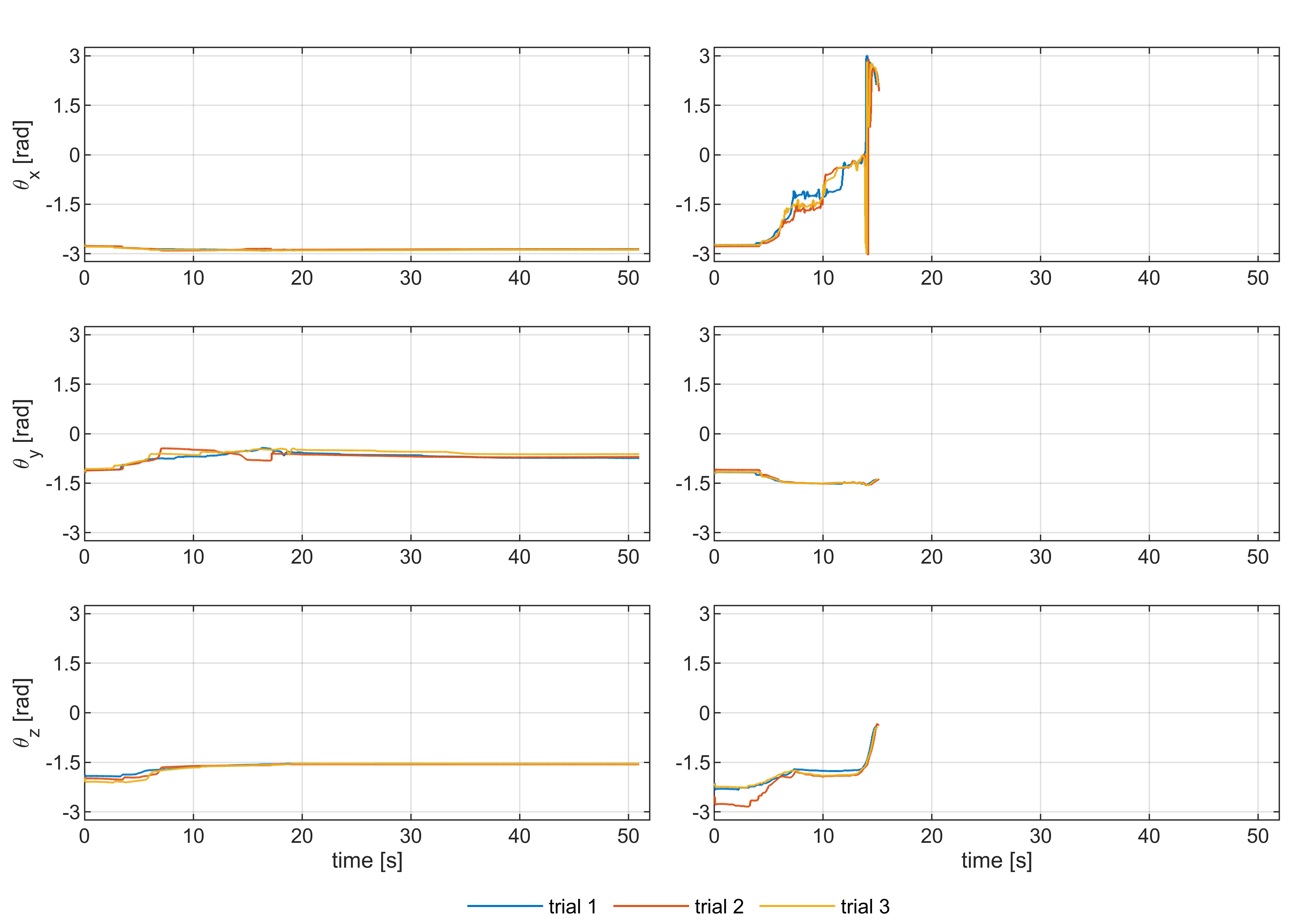}}
	\caption{Evaluating the repeatability of the experimental setup: \textbf{(A)} linear and \textbf{(B)} angular displacements resulting from testing the RRR SoftFoot three times on the flat ground, both for a stable (left column) and an unstable (right column) point.}
	\label{repeatability}
\end{figure*}

Benchtop testing through a robotic arm is chosen because it ensures versatility, repeatability, and reproducibility of the experimental session, which are fundamental when testing many configurations of feet and uneven ground profiles. 
The experimental setup is validated before starting the experiments. Repeatability was investigated by repeating three times a trial with the RRR configuration of the SoftFoot on a flat ground, for both a stable and an unstable point (see next subsection for the definitions of stability and instability). Results are shown in Figure \ref{repeatability} in terms of linear displacements and rotations for both points. Despite the expected worse result for the unstable point, results confirm a good overall setup repeatability.

\subsubsection{Procedure}\label{sssec:num2}

\begin{figure*}[ht]
	\centering \includegraphics[width=1\linewidth]{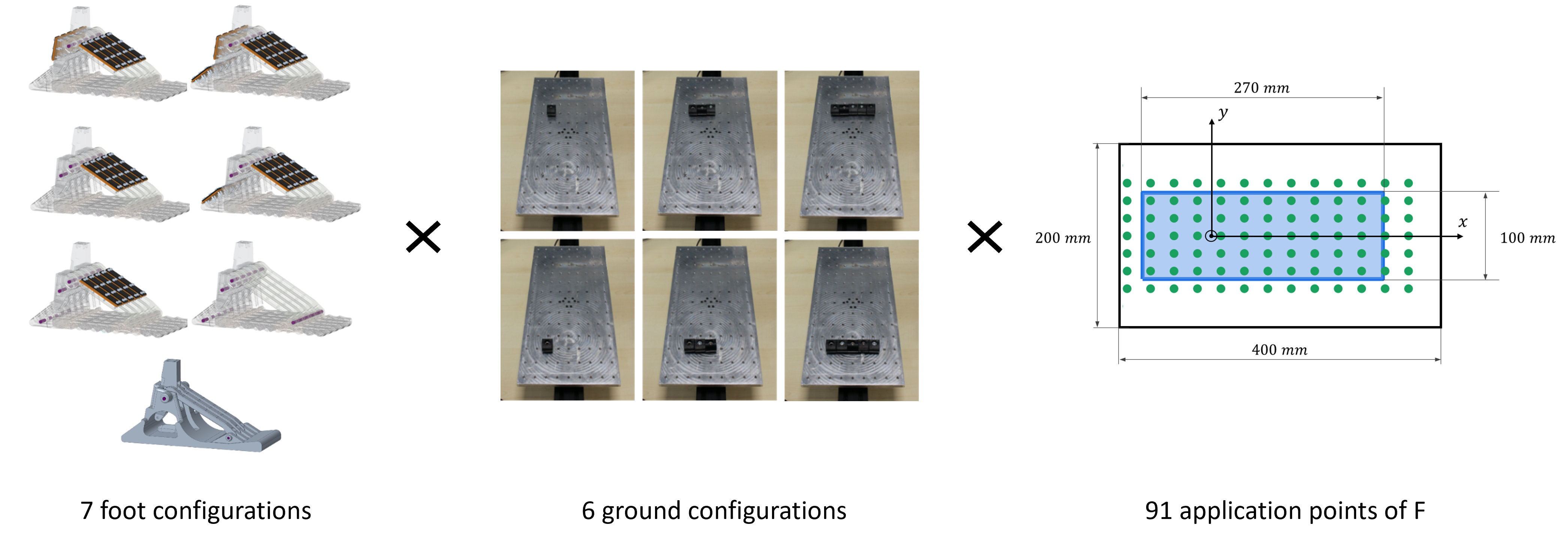}
	\caption{Experimental protocol: on the left, the five configurations of the SoftFoot 3D, the SoftFoot described in \cite{piazza2016, piazza2024}, and a completely rigid foot; in the center, the 6 different plate configurations with obstacles of various sizes simulating different uneven grounds (respectively, from left to right, size S, M, and L obstacles, placed under the heel - top row - and the forefoot - bottom row); on the right, the light blue rectangle represents the unloaded foot base, with the foot oriented so as to have the toes in the positive direction of the x-axis, the origin of the reference frame corresponds to the ankle projection (specifically, the projection of the proximal extremity of the central rear arch, which allows the connection with the user's leg), and the green dots to the application points of the resulting force (sum of the vertical force applied by the robotic arm and the foot weight).}
	\label{exp}
\end{figure*}

The six SoftFoot configurations and the completely rigid foot (Figure \ref{exp}, on the left) are tested on uneven grounds made of three different sizes of obstacles (Figure \ref{exp}, in the center): \textit{size S} refers to a single cylindrical obstacle; \textit{size M} to three cylindrical obstacles; \textit{size L} to five cylindrical obstacles placed symmetrically with respect to the longitudinal axis of the plate. Experiments are conducted firstly with the obstacles placed under the heel and secondly under the forefoot. These are, indeed, the two areas of the foot more stressed during two critical events of the gait cycle, i.e. respectively heel strike and toe-off, when foot instability due to an uneven ground can jeopardise the whole user's stability resulting in a fall. As a result, six different ground profiles are tested (Figure \ref{exp}, in the center).

The robotic arm is controlled in order to apply on the foot in contact with the plate, for each ground configuration, a 2 kg weight ($m_{load}$). The point of application of this weight is constant with respect to the plate during each trial, while it is moved in the xy-plane between trials at the 91 threaded holes in the aluminum plate. Given the foot mass ($m_{foot}$) of approximately 1 kg (1.115 kg, except 1.13 kg of the rigid foot), the z-component of the resulting force, perpendicular to the ground, is about 30 N (i.e. $(m_{load}+m_{foot})g$) for all the application points of the force - measured by the upper ATI and reported with respect to the reference frame of the plate. The application points of the resulting force (green dots in Figure \ref{exp}, on the right) lie within the unloaded foot base (i.e. the light blue rectangle in Figure \ref{exp}, on the right), or in the area very close to its perimeter to account also for possible (indeed observed) foot deformation. Those points placed outside the unloaded foot base are referred to as \textit{external points} in the rest of the paper.

The specific $m_{load}$ is chosen considering the robotic arm payload (3 kg) and the foot mass ($m_{foot}$). The force is slowly applied on the seven foot configurations through the robotic arm to simulate quasi-static conditions, neglecting inertial terms and simplifying the calculations. Given the 7 foot configurations to be tested on 6 different grounds and loaded in 91 application points, a total of 3822 trials are conducted, lasting more than 128 hours (excluding the setup time).

Each trial is performed according to the four phases reported in Table \ref{phases}, and implementing a specific control for the robotic arm in each phase, either impedance (to obtain a kind of position control) or force control. A calibration is performed at the beginning to record the relative position of the reference frames of the plate and of the robotic arm base (by using the robot pose and the setup dimensions).
\begin{table}[h!]
	\caption{Description of the four phases characterizing each experimental trial with the corresponding control implemented in the robotic arm (\textit{Imp} and \textit{Force} respectively for an impedance and a force control).\vspace{3mm}}
	\label{phases}
\centering

\begin{tabular}{|p{70pt} |  p{330pt}  | P{22pt} | P{24pt}|}
\hline
\textbf{Phase} & \centering\arraybackslash{\textbf{Description}} & \centering\arraybackslash{\textbf{Imp}} & \centering\arraybackslash{\textbf{Force}}\\
\hline
\hline
\savecellbox{\textbf{Homing}} & \savecellbox{The foot is kept in the \textit{home} pose, i.e. a vertical displacement of about 100 mm of the foot center of mass (COM) with respect to the ground.} & 
\savecellbox{$\bullet$} & \savecellbox{} \\ 
[-\rowheight]
\printcellmiddle & \printcellmiddle & \printcellmiddle & \printcellmiddle\\
\hline
\savecellbox{\textbf{Reaching}} & \savecellbox{The foot is moved towards the aluminum plate in the vertical direction (average velocity about 2 mm/s) until it touches the ground, corresponding to an approximately 11 N (i.e. $m_{foot}g$) force measured at the plate.} & \savecellbox{$\bullet$} & \savecellbox{} \\
[-\rowheight]
\printcellmiddle & \printcellmiddle & \printcellmiddle & \printcellmiddle \\
\hline
\savecellbox{\textbf{Pushing}} & \savecellbox{A vertical force is applied on the foot placed on the plate through the robotic arm. The force wrench is slowly and linearly increased until \textit{stability} or \textit{instability} occurs: in the first case, the component of the resulting force perpendicular to the ground reaches about 30 N, whereas in the second case, the foot falls down.} & \savecellbox{} & \savecellbox{$\bullet$}\\ 
[-\rowheight]
\printcellmiddle & \printcellmiddle & \printcellmiddle & \printcellmiddle \\
\hline 
\savecellbox{\textbf{Withdrawing}} & \savecellbox{The force is reduced to zero, and the robotic arm is moved back to the \textit{home} pose, ready for the next trial.} & \savecellbox{$\bullet$} & \savecellbox{}\\ 
[-\rowheight]
\printcellmiddle & \printcellmiddle & \printcellmiddle & \printcellmiddle \\
\hline
\end{tabular} 
\end{table}

During the pushing phase, two scenarios might happen: 
\begin{enumerate}
\item\ The resulting force measured by the upper ATI reaches 30 N in the vertical direction, while the foot reaches a steady state condition, which is kept until the duration of the pushing phase reaches 40 s, before returning to the \textit{home} pose (see definition in Table \ref{phases}). The corresponding application point of the force is defined as \textit{stable}. 
\item\ The resulting force measured by the upper ATI does not reach 30 N, because the foot falls down. Specifically, the EE reaches a displacement $\geqslant$ 100 mm and/or a rotation $\geqslant$ 45° around at least one axis with respect to the undisturbed pose (i.e. the foot lying on the plate in equilibrium without applied forces). In this case, the corresponding point is defined as \textit{unstable}.
\end{enumerate}

The thresholds set for displacements and rotations are heuristically determined by looking at the foot behavior during initial testing. They proved to be appropriate for discriminating instability. All the tested feet, indeed, after surpassing those thresholds, completely lose contact with the ground. So it is not possible to register cases of false instability, i.e. after an initial slipping the foot assumes a configuration that conforms to the uneven ground.

\begin{figure}[h!]
	\centering
	{\includegraphics[width=0.5\textwidth]{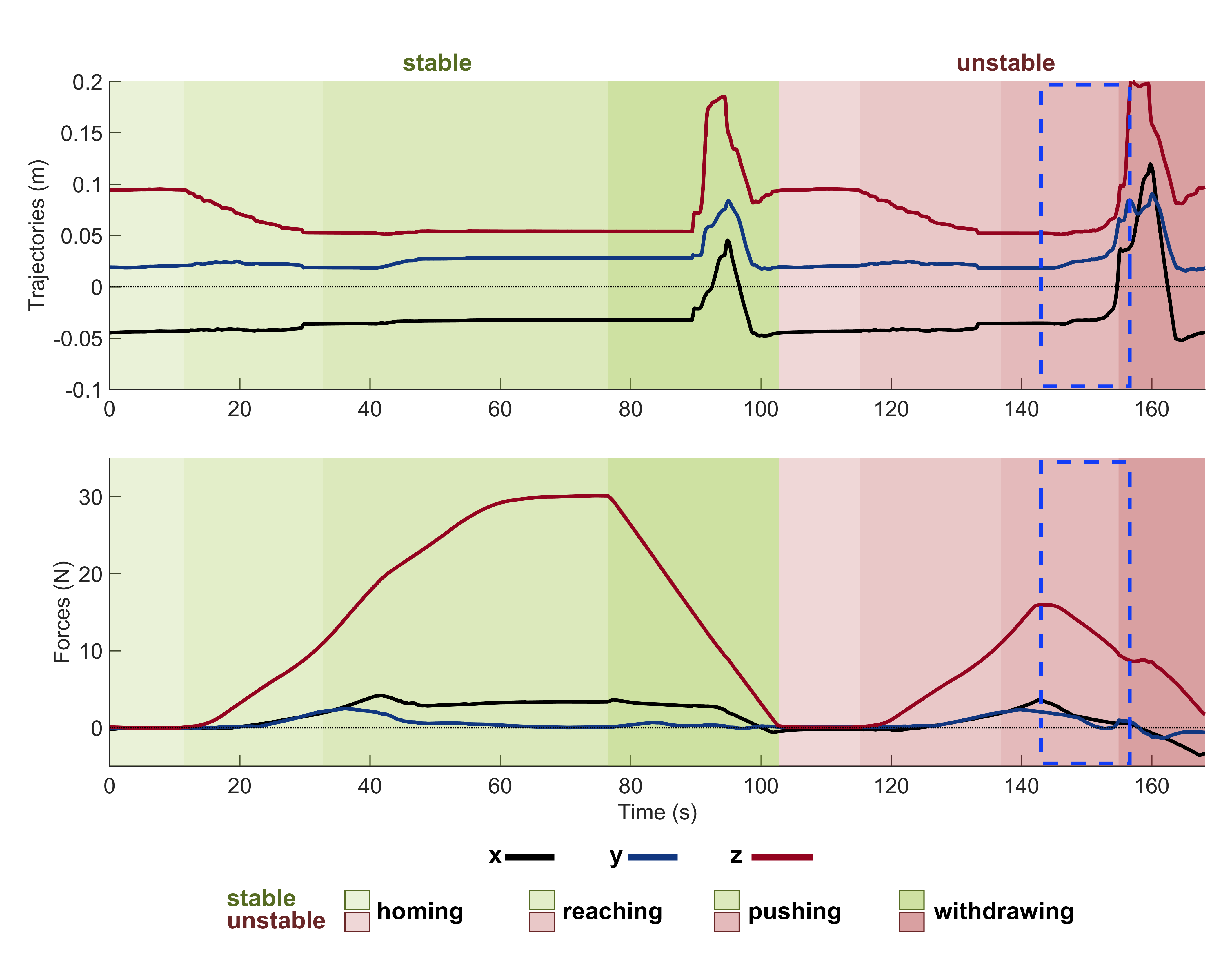}} 
	\caption{Results of two consecutive experimental trials conducted with a size S obstacle placed under the forefoot of the RRR SoftFoot. The top plot displays the foot (i.e. its COM) trajectories with respect to the plate reference frame; the bottom plot the force measured by the upper ATI and plotted with respect to the plate, resulting by summing the vertical force applied by the robotic arm and the foot weight. The two background colors indicate respectively the stability - green shades - and instability - light red shades - of the trials. The different shades of each color denote the four phases characterizing each trial. The dashed blue rectangular profiles highlight the time window in which instability occurs.}
	\label{st_unst_point}
\end{figure}

By way of example, Figure \ref{st_unst_point} shows the results of two consecutive trials with a size S obstacle placed under the forefoot of the RRR SoftFoot. In a trial that turns out to be unstable, foot trajectories quickly vary during the pushing phase because of the large EE linear and/or angular displacement (see the dashed blue rectangle in Figure \ref{st_unst_point}), whereas they reach a steady state for a stable point, with the z-component of the resulting force reaching approximately 30 N. The largest displacement of the feet during testing occurs along the z-axis, while the trajectories along the x- and y-axes are almost constant during the homing and reaching phases, show small changes during pushing, and peaks during withdrawing caused by the larger movements of the foot while returning to the home pose. For a better understanding of the experimental procedure, a video of two consecutive trials conducted with one of the SoftFoot configurations and a \textit{size L} obstacle placed under the forefoot is available as a supplementary video.

\section{Experimental results} \label{experiments}

The performance of different SoftFoot on different grounds is compared by considering the number of points that are found to be stable among all those tested on the plate.

Table \ref{st_unst_table} displays in brackets in the first column the number of stable points ($n$) for each foot-obstacle combination, out of all the 91 tested points. Moreover, the percentage of stable points obtained with each SoftFoot (SF, i.e. configurations (A)-(F) in Figures \ref{conf} and \ref{dofs}) with respect to those obtained with the rigid foot (RF, i.e. the reference case representing the 100\%, Figure \ref{dofs}G) is calculated as follows (and displayed in the first column ($\%$) of Table \ref{st_unst_table}):
\[\% = \frac{n \text{ stable points SF}}{n \text{ stable points RF}} \cdot 100\] 

Table \ref{st_unst_table} displays also in the second column the number of external stable points ($n_E$), which are those stable points placed outside the unloaded foot base, i.e. outside the light blue rectangle in Figure \ref{exp}. Theassociated percentage of external stable points with respect to the total number of stable points of the same foot is also showed ($\%_E$), and it is calculated as follows:
\[\%_E = \frac{n\ \text{ext stable points SF}}{n \text{ stable points SF}} \cdot 100\]

The KRK SoftFoot exhibits the largest number of stable points when the obstacle is placed under the forefoot. Together with the KRF, it also ensures overall better stability when the obstacle is hit with the heel, albeit all the SoftFoot configurations result to be less stable than the rigid foot in this case, except for the small obstacle. With a large obstacle under the heel, three SoftFoot configurations, i.e. the KRF, KRK, and RRR, show the same performance in terms of stable points.

The KRF, KRK, and RRR SoftFoot configurations perform poorly in terms of their capability to extend their stability area in points that were not originally included in the unloaded foot base (see $\%_E$ and $n_{E}$). In some cases, indeed, they have the same performance of the rigid foot, which cannot extend (the corresponding number of external points and percentage are null) since its rigid structure limits its stability area (i.e. the convex hull of contact points) to the unloaded foot area. In contrast, the KKF, KKK, and KRR SoftFoot configurations exhibit an overall better capability to extend their stability area.

Figure \ref{stunst}A gives a visual insight into the feet stability when tested on uneven ground, respectively with a size S, M, and L obstacle, under the heel and the toes. The dots, representing the 91 application points of the vertical force, are shown in green when stable and in red when unstable (see definition in subsection \ref{sssec:num2}). The RRR was tested too for comparison purposes, and the rigid foot was kept as a reference.

Those SoftFoot configurations that turned out to be the most stable during the aforementioned experimental session (results in Figure \ref{stunst}A), i.e. KRF and KRK, were tested with a set of stiffer coil springs (1.7 N/mm, i.e. one order of magnitude higher than the first tested set). Results are presented in the same Table \ref{st_unst_table} (last four rows) and visually in Figure \ref{stunst}B. The same trend can be observed: the SoftFoot outperforms the rigid foot anytime obstacles are placed under the forefoot, as well as it shows equal or better performance when a small obstacle is hit with the heel (except for the RRR configuration). 
\begin{table*}[hb!]
	\centering
	\caption{For each foot-obstacle combination, in the first column, number of stable points ($n$) and corresponding percentage ($\%$) with respect to the number of stable points of the rigid foot (representing the 100\%). Cell color legend: red if $\% < 90\%$, orange if $90\% \leq \% < 110\%$, yellow if $110\% \leq \% < 130\%$, green if $\% > 130\%$. In the second column, number of external stable points ($n_E$) and corresponding percentage ($\%_E$) with respect to the total number of stable points of that combination. The average number of stable points for each foot is displayed in the last column.\vspace{3mm}}
 	\label{st_unst_table}
	\begin{adjustbox}{width=1\textwidth}
\small
	\begin{tabular}{|c|c c|c c|c c|c c|c c|c c|c|}
		\hline
		 & \multicolumn{6}{c|}{\textbf{Heel}} & \multicolumn{6}{c|}{\textbf{Toes}} &\\
		 & \multicolumn{2}{c}{\textbf{S}} & \multicolumn{2}{c}{\textbf{M}}& \multicolumn{2}{c|}{\textbf{L}} & \multicolumn{2}{c}{\textbf{S}} & \multicolumn{2}{c}{\textbf{M}} & \multicolumn{2}{c|}{\textbf{L}} & $n_{average}$\\
		& \% ($n$)
		& $\%_E$ ($n_{E}$) 
		& \% ($n$)
		& $\%_E$ ($n_{E}$) 
		& \% ($n$)
		& $\%_E$ ($n_{E}$) 
		& \% ($n$)
		& $\%_E$ ($n_{E}$) 
		& \% ($n$)
		& $\%_E$ ($n_{E}$) 
		& \% ($n$)
		& $\%_E$ ($n_{E}$)
  & 
		 \\
		 \hline \hline
		\textbf{Rigid}	&
			\cellcolor{Dandelion}\textbf{100 (38)} 
						&0 (0)&
			\cellcolor{Dandelion}\textbf{100 (36)} 
						&0 (0)&
			\cellcolor{Dandelion}\textbf{100 (47)}
						&0 (0)&
			\cellcolor{Dandelion}\textbf{100 (33)}
			        	&0 (0) &
			\cellcolor{Dandelion}\textbf{100 (27)} 	
					&	0 (0) &
			\cellcolor{Dandelion}\textbf{100 (35)} 
						&0 (0)
      & \textbf{36}\\

		\cellcolor{gray!20}\textbf{KKF}	& 
		\cellcolor{RedOrange}\textbf{87 (33)} 	
			&	\cellcolor{gray!20}9 (3) &
		\cellcolor{RedOrange}\textbf{89 (32)}	
			&	\cellcolor{gray!20}16 (5)&
		\cellcolor{RedOrange}\textbf{85 (40)} 	
			& 	\cellcolor{gray!20}8 (3)&
		\cellcolor{Yellow}\textbf{115 (38)}  
			&   \cellcolor{gray!20}11 (4) &
		\cellcolor{Yellow}\textbf{126	(34)} 	
			&	\cellcolor{gray!20}6 (2)  &
		\cellcolor{Dandelion}\textbf{91 (32)} 	
			& 	\cellcolor{gray!20}3 (1)
   & \cellcolor{gray!20}\textbf{34.8} \\

			\textbf{KKK}	&
			\cellcolor{Dandelion}\textbf{108 (41)}
					&10 (4) &
					\cellcolor{RedOrange}\textbf{81 (29)}
					&10 (3)&
					\cellcolor{RedOrange}\textbf{77 (36)}
					&	6 (2)&
					\cellcolor{Yellow}\textbf{115 (38)}
					&3 (1) &
					\cellcolor{Yellow}\textbf{126 (34)} 
					&9 (3)  &
					\cellcolor{Dandelion}\textbf{100 (35)}
					&	9 (3)
     & \textbf{35.5}\\

			\cellcolor{gray!20}\textbf{KRF} &
			\cellcolor{Dandelion}\textbf{100 (38)}	
			&	\cellcolor{gray!20}3 (1)
			&	\cellcolor{Dandelion}\textbf{94 (34)}
			&	\cellcolor{gray!20}3 (1)
			&	\cellcolor{RedOrange}\textbf{89 (42)}
			&	\cellcolor{gray!20}0 (0) &
			\cellcolor{Yellow}\textbf{118 (39)}
			& \cellcolor{gray!20}5 (2)
			&	\cellcolor{Yellow}\textbf{122 (33)}
			&	\cellcolor{gray!20}0 (0)
			&	\cellcolor{Yellow}\textbf{114 (40)}
			&	\cellcolor{gray!20}0 (0)
   & \cellcolor{gray!20}\textbf{37.7}\\	
			
		\textbf{KRK}& 
			\cellcolor{Dandelion}\textbf{108 (41)}
			&	0 (0)
			&	\cellcolor{RedOrange}\textbf{86	(31)}  
			&	3 (1)
			&	\cellcolor{RedOrange}\textbf{89 (42)}   
			&	0 (0)
			&   \cellcolor{Yellow}\textbf{124 (41)}
			&	2 (1)
			&	\cellcolor{YellowGreen}\textbf{148 (40)}  
			&	0 (0)
			&	\cellcolor{Yellow}\textbf{120 (42)}  
			&	5 (2)
   & \textbf{39.5}\\	

			\cellcolor{gray!20}\textbf{KRR}& 
			\cellcolor{Dandelion}\textbf{103 (39)}
			&	\cellcolor{gray!20}5 (2)
			&	\cellcolor{RedOrange}\textbf{83 (30)}    
			&	\cellcolor{gray!20}3 (1)
			&	\cellcolor{RedOrange}\textbf{79	(37)}   
			&	\cellcolor{gray!20}8 (3)
			&   \cellcolor{Dandelion}\textbf{109 (36)}   
			&	\cellcolor{gray!20}3 (1) 
			&   \cellcolor{Yellow}\textbf{130 (35)}   
			&	\cellcolor{gray!20}6 (2)
			&	\cellcolor{Dandelion}\textbf{100 (35)} 
			&	\cellcolor{gray!20}9 (3)
   &\cellcolor{gray!20}\textbf{35.3}\\	

			\textbf{RRR}& 
			\cellcolor{Dandelion}\textbf{100 (38)}  
			&	3 (1)
			&	\cellcolor{RedOrange}\textbf{67 (24)} 
			&	0 (0)
			&	\cellcolor{RedOrange}\textbf{89	(42)}
			&	0 (0)
			&   \cellcolor{Yellow}\textbf{121 (40)}
			&	3 (1)
			&   \cellcolor{Yellow}\textbf{122 (33)}
			&	0 (0)
			&   \cellcolor{Yellow}\textbf{114 (40)}
			&	0 (0)
   & \textbf{36.2}\\	

		\hline
  \multicolumn{14}{|c|}{\cellcolor{gray!50}\textbf{with a stiffer set of coil springs}}\\
  \hline
		\textbf{Rigid}	&
		\cellcolor{Dandelion}\textbf{100 (38)}
		&0 (0)&
		\cellcolor{Dandelion}\textbf{100 (36)}
		&0 (0)&
		\cellcolor{Dandelion}\textbf{100 (47)} 
		&0 (0)&
		\cellcolor{Dandelion}\textbf{100 (33)} 
		&0 (0) &
		\cellcolor{Dandelion}\textbf{100 (27)} 
		&	0 (0) &
		\cellcolor{Dandelion}\textbf{100 (35)} 
		&0 (0) & \textbf{36}\\

		\cellcolor{gray!20}\textbf{KRF} 
        &	\cellcolor{Dandelion}\textbf{100 (38)}
		&	\cellcolor{gray!20}0 (0)
		&	\cellcolor{RedOrange}\textbf{81 (29)}
		&	\cellcolor{gray!20}0 (0)
		&	\cellcolor{RedOrange}\textbf{89 (42)}
		&	\cellcolor{gray!20}0 (0)
		&   \cellcolor{YellowGreen}\textbf{145 (48)} 	
		&	\cellcolor{gray!20}10 (5)
		&	\cellcolor{Yellow}\textbf{122 (33)}  
		&	\cellcolor{gray!20}0 (0)
		&	\cellcolor{Yellow}\textbf{114	(40)}  
		&	\cellcolor{gray!20}0 (0) 
        & \cellcolor{gray!20}\textbf{38.3}\\	
		
		\textbf{KRK}& 
		\cellcolor{Yellow}\textbf{113 (43)}
		&	0 (0)
		&	\cellcolor{Dandelion}\textbf{97 (35)}
		&	0 (0)
		&	\cellcolor{Dandelion}\textbf{94 (44)}
		&	5 (2)
		&   \cellcolor{Yellow}\textbf{127 (42)} 
		&	0 (0)
		&	\cellcolor{YellowGreen}\textbf{159 (43)}
		&	0 (0)
		&	\cellcolor{YellowGreen}\textbf{140 (49)}
		&	0 (0) & \textbf{42.7}\\	
		
		\cellcolor{gray!20}\textbf{RRR}& 
		\cellcolor{RedOrange}\textbf{89 (34)}
		&	\cellcolor{gray!20}0 (0)
		&	\cellcolor{Dandelion}\textbf{92 (33)} 
		&	\cellcolor{gray!20}0 (0)
		&	\cellcolor{Dandelion}\textbf{96 (45)}
		&	\cellcolor{gray!20}0 (0)
		&   \cellcolor{Yellow}\textbf{118 (39)} 
		&	\cellcolor{gray!20}5 (2)
		&\cellcolor{YellowGreen}\textbf{144 (39)}
		&	\cellcolor{gray!20}0 (0)
		&\cellcolor{YellowGreen}\textbf{137 (48)}
		&	\cellcolor{gray!20}17 (8)  
        & \cellcolor{gray!20}\textbf{39.7}\\	
\hline
	\end{tabular}
	\end{adjustbox}
\end{table*}

The increased heel stiffness leads to a greater number of stable points for the KRK SoftFoot in all cases (Table \ref{st_unst_table} and Figure \ref{stunst}B). The KRF configuration shows the same or an increased number of stable points when the obstacle is under the toes, but the same or reduced number when the obstacle is under the heel. The RRR configuration exhibits better stability only when medium and large obstacles are considered.

The bar chart in Figure \ref{barplot} gives an overview of the percentages reported in Table \ref{st_unst_table} for those SoftFoot 3D configurations tested with both the softer and the stiffer set of springs (i.e. KRF and KRK), together with the RRR SoftFoot and the rigid foot for comparison.
\section{Discussion} \label{discussion}

The superior performance of the RRR configuration with respect to the rigid foot in terms of number of stable points when stepping on obstacles under the forefoot confirms that the SoftFoot design and, thus, the use of basic modules with intrinsic longitudinal adaptability, is effective in obstacle negotiation. The KRK and KRF SoftFoot configurations ensure the greatest stability on average across all ground profiles. They are both characterized by an elastic connection at the frontal arches, rigidly connected rear arches, and a not-rigid connection across the heels. Thus, their performance suggests that the introduction of the transverse arch in the SoftFoot is advantageous in coping with unevenness, especially when placed under the forefoot: it leads to improved obstacle negotiation and overall stability compared to a traditional rigid foot and to the SoftFoot featuring adaptability only in the sagittal plane.

An elastic connection across the heels allows for the best performance in filtering small obstacles under the heel, contributing to enhancing frontal adaptability. On the contrary, bigger obstacles at the heel are more difficulty filtered, jeopardizing the stability of the analyzed adaptive feet, with a free connection at the heel that seems to be helpful in case of medium obstacles.

\begin{figure}[hb!]
	\centering
	\includegraphics[width=1\linewidth]{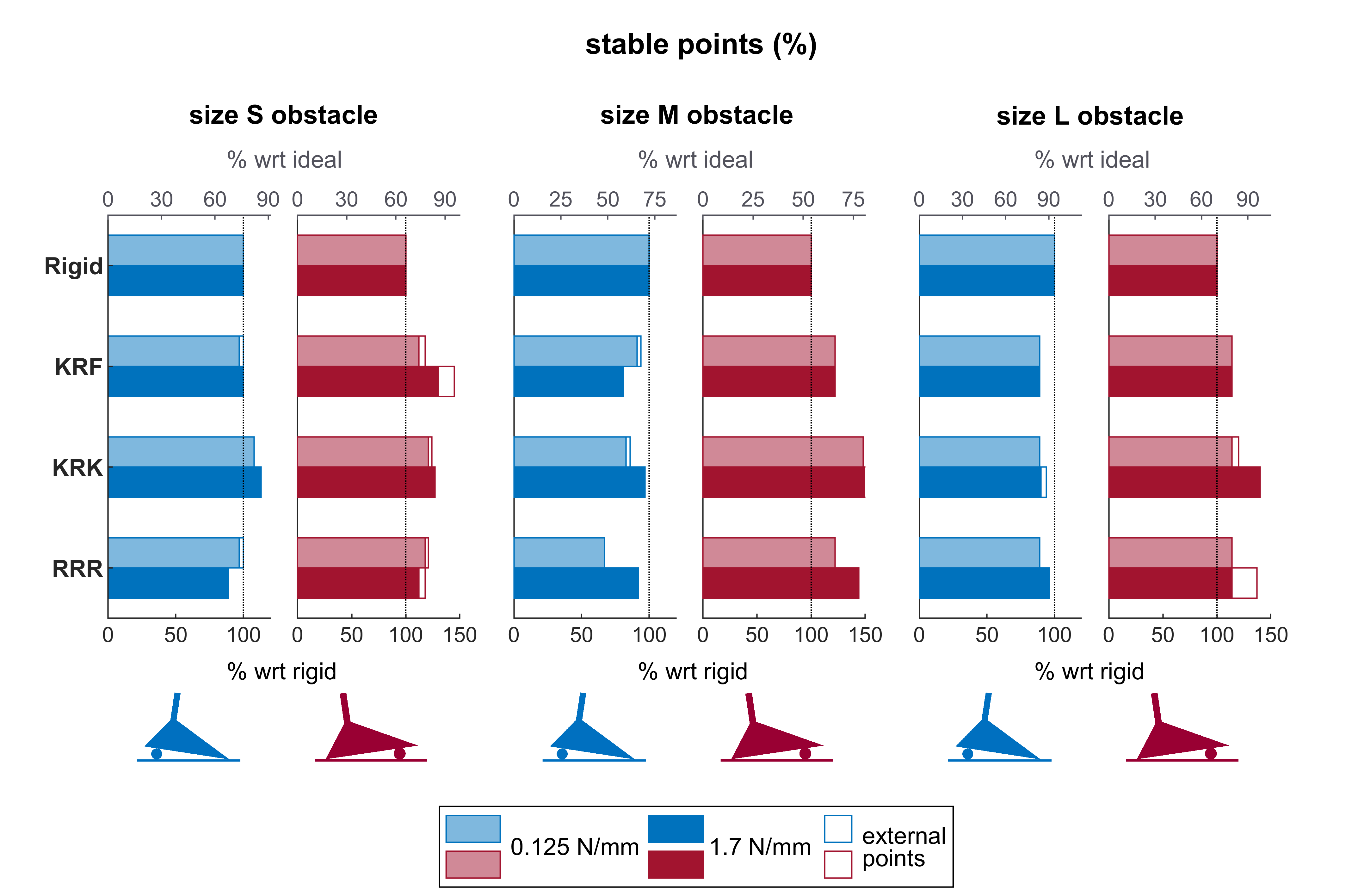}
	\caption{Bars showing the percentage of stable points for the rigid foot and the three SoftFoot configurations (i.e. KRF, KRK, RRR) tested with both the softer (lighter shades) and the stiffer (darker shades) set of springs on the three ground profiles (i.e. the three obstacle sizes (S, M, L)), and with obstacles placed under the heel (blue bars) and the forefoot (red bars). The lower x-axis describes the percentage of stable points for each SoftFoot with respect to those of the rigid foot (representing 100\%), while the upper x-axis (in grey) describes the percentage of stable points for each foot with respect to the ideal number of stable points, which is the total number of points included in the foot base when the unloaded foot is on the even ground, i.e. 50 points (representing 100\%). The non-filled parts of the bars represent the percentage of "external" stable points, i.e. placed outside the unloaded foot base.}
	\label{barplot}
\end{figure}

\begin{figure}[ht!]
	\centering
 \subfloat[]
	{\includegraphics[width=0.96\textwidth]{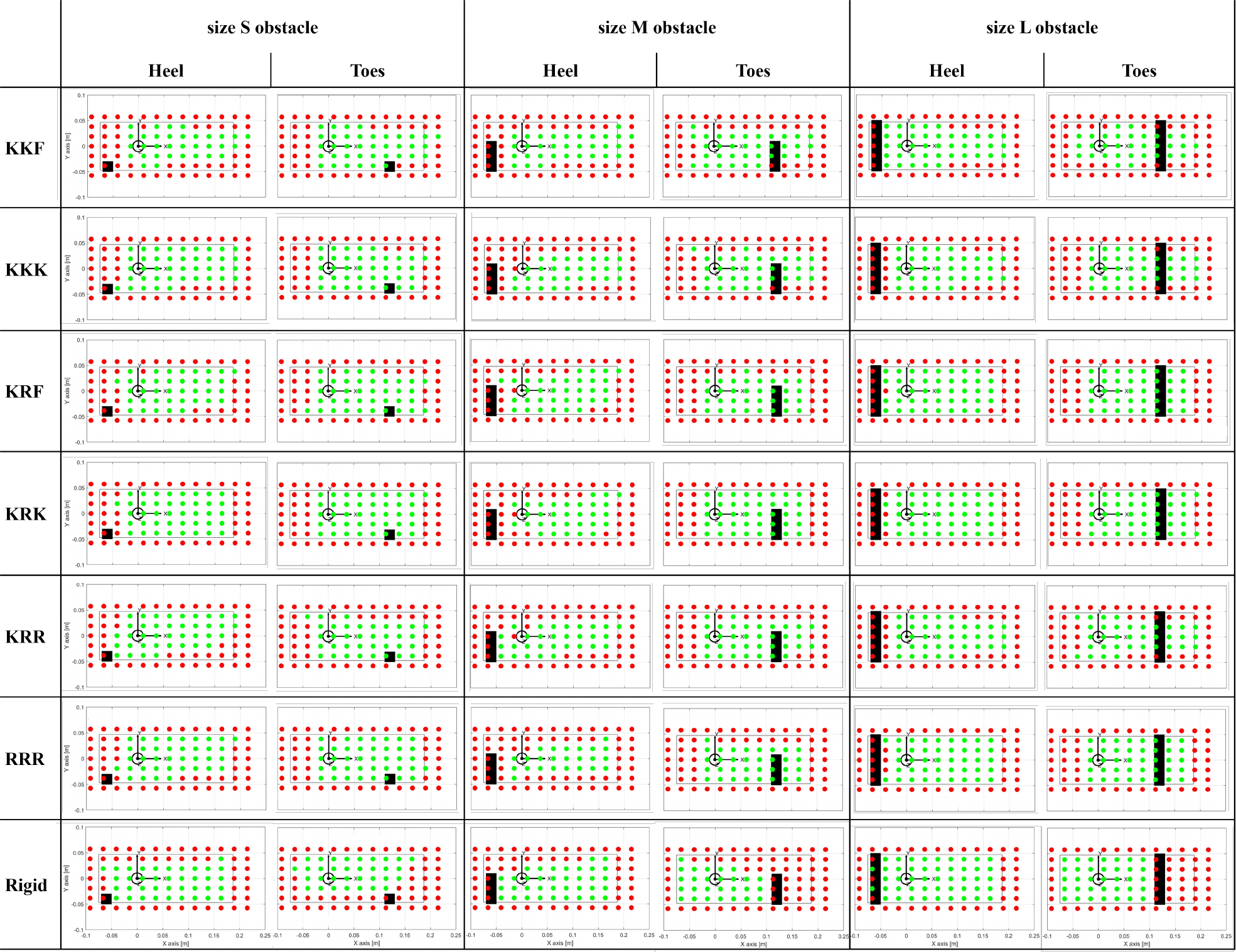}}\\
 \subfloat[]
 {\includegraphics[width=0.96\textwidth]{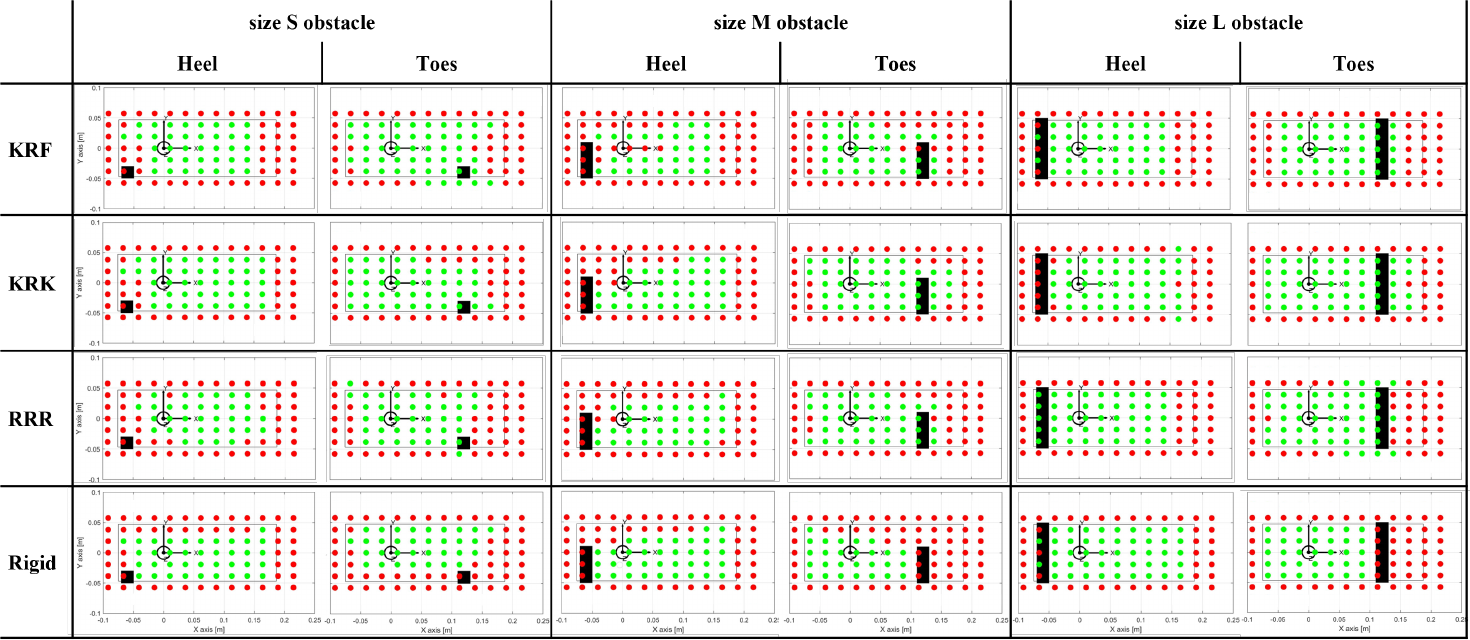}}
	\caption{Stable (green dots) and unstable (red dots) points for different foot-obstacle combinations. The black inner rectangular profile represents the area of the unloaded tested foot (with the heel placed by the negative values of the x-axis); the origin of the reference frame corresponds to the ankle projection; the black filled rectangle to the area of the obstacles. \textbf{(A)} The tested feet are the six SoftFoot configurations, and the rigid foot. \textbf{(B)} The tested feet are the three SoftFoot configurations with stiffer springs (springs constant 1.7 N/mm), and the rigid foot.}
	\label{stunst}
\end{figure}

A rigid connection across rear arches, like in the KRF and KRK configurations, seems to improve stability when bumping into obstacles under both the forefoot and the hindfoot, perhaps by keeping the foot structure more compact. Conversely, an elastic connection allows for greater deformation of the foot structure when loaded on unevenness, fostering the extension of the foot stability area beyond the unloaded foot base (see larger number of external stable points). This means a larger contact surface, which is a desirable feature in biped robots' feet, benefitting their balance and increasing their dynamic range by enlarging the support area on which the robot-ground interaction forces act.

Interestingly, as previously discussed, but also from a visual analysis of Figure \ref{stunst}A, the heel turns out to be the weakest area of the SoftFoot, regardless of whether the obstacle is placed under the heel or the toes. When loaded in the heel area, indeed, the SoftFoot results to be often unstable (i.e. see red dots). We supposed this behavior to be related to the stiffness of the springs adopted to connect the rear arches to the sole: they are likely not sufficiently stiff to prevent the backward closure of the rear arch towards the heel, especially when the force is applied on the back area of the foot, resulting in instability.

For this reason, an additional experimental session was conducted: testing was repeated for those configurations that proved to be the most stable during the first experimental session, namely the KRK and KRF SoftFoot, together with the RRR for comparison, using stiffer springs. This change led to an increased number of stable points on average for the three SoftFoot configurations (Table \ref{st_unst_table}), but not to an overall larger number of external stable points (i.e. a stiffer hindfoot does not foster the extension of the foot stability area in points outside the unloaded foot area). The KRK SoftFoot configuration performs better than the others on average. It outperforms the rigid foot and the SoftFoot with only longitudinal adaptability for obstacles of any size placed under the toes, as well as for small obstacles under the heel.

This trend for both the softer and the stiffer set of springs emerges in Figure \ref{barplot} too. The same figure also shows how the KRK configuration is, on the whole, the closest to the ideal case, which is given by a number of stable points equal to the total number of points covered by the unloaded foot sole on even ground (i.e. 50 points).

By visually inspecting Figure \ref{stunst}, a trend can be observed: increasing springs stiffness improves heel stability at the expense of the forefoot stability. Indeed, the number of stable points in the heel area (see green dots, especially when the obstacle is placed under the forefoot) increases, while the stable points under the forefoot decrease (see increased number of red dots at the toes level).  Likely, the stiffening of the rear part of the foot affects the overall foot deformation capability and, thus, its stability when loaded at the toes level. Future work will include a sensitivity study to further investigate the relationship between springs stiffness and heel-forefoot stability, considering sets of springs whose stiffness ranges between the two spring constants considered in this study (i.e. 0.125 N/mm and 1.7 N/mm). A threshold could be found to improve the stability of the SoftFoot when loaded in the heel area, while preserving at most its stability when loaded at the forefoot, as well as its ability to extend its stability area in points outside the unloaded foot area.

\subsection{Further considerations}
In this study, we connected each link and joint to their analogs in the adjacent modules, as well as we connected the five modules in the same way. Future investigations we will re-examine the preliminary assumptions and the considerations made to simplify the design process. By considering different connections between the five modules, for instance, we might discover additional sensible combinations and, thus, further designs to be analyzed.

In addition, changes in the performance of the SoftFoot configurations in terms of stable and unstable points might occur when the elastic connection among 2D modules is implemented differently. This aspect is left for future work. A sensitivity study will be performed to investigate the changes in foot stability when, for example, the shore hardness and the thickness of the rubber sheets vary. Likely, an appropriate range will emerge.

Note that a Boolean approach is used to evaluate stability at each point of application of the resulting force, rather than investigating the level of stability. The former represents, indeed, an approach close to reality, where the primary concern, for both biped robots and human beings, is to be stable and not to fall. However, the level of stability achieved is an aspect that may be worth investigating in the future.

This study paves the way for the engineering of a soft foot for prosthetic applications. In fact, commercial prosthetic feet, as well as state-of-the-art research prototypes, are usually characterized by flat and stiff soles that hinder ground adaptation in case of obstacles, similar to robotic feet. This introduces instability with the associated risk of falling, and leads to compensatory mechanisms, additional stresses on the residual limb, and physical injuries in the long term \citep{weerakkody2017}. Therefore, a prosthetic foot capable of achieving compliant interaction with the environment - both in the sagittal and frontal planes - while preserving the stability of the user, could make a difference in the lives of people using lower-limb prostheses.
\section{Conclusions}
This paper investigates the role of the longitudinal and transverse arches in a class of anthropomorphic soft robotic foot inspired by the architecture of the SoftFoot \citep{piazza2016, piazza2024}, for enhanced performance on unstructured terrains.

The class of feet we analyzed was obtained by considering all possible connections between five parallel basic modules characterized by intrinsic longitudinal adaptability. This makes the modules able to move relative to each other to obtain adaptability in the frontal plane. Through a series of assumptions and considerations, we carefully excluded those configurations that would present functional, kinematic, and implementational limitations, narrowing down our research to five configurations. They were implemented and tested experimentally, together with the SoftFoot described in \cite{piazza2016, piazza2024} and a rigid flat foot, for comparison.

Results prove that the design of a compliant foot embodying two arches, as those of the human foot, is promising for robotic applications. Indeed, those soft robotic feet featuring intrinsic adaptability in both the sagittal and frontal planes exhibited a clear advantage in obstacle negotiation, especially when obstacles were hit with the forefoot, as this is the most compliant part of those soft feet. They exceeded the performance of both a rigid flat foot and a soft foot featuring only longitudinal adaptability. Specifically, by elastically connecting the frontal arches, and either by using an elastic connection across the heels or by letting them free (corresponding respectively to a KRK and a KRF configuration), it was possible to achieve greater stability (i.e. a larger number of stable points) on most of the ground profiles, filtering unevenness of different dimensions.

A sensitivity study will be conducted in future work to find the optimal stiffness for the heel springs providing the greatest stability when the foot is loaded both at the hindfoot and forefoot level. Similarly, the assumptions and considerations made at the beginning of the study will be revised. Furthermore, the SoftFoot configurations analyzed in this paper will be evaluated also with biped robots, to thoroughly investigate the connection between foot adaptability to the ground and overall robots' stability in real-world environments.

Lastly, in addition to the robotics community, these findings might greatly benefit the prosthetic sector too. 
The need for prosthetic feet that allow prosthetic users to walk safely on uneven terrains by compliantly adapting in both the longitudinal and frontal planes is, indeed, still actual.

\section*{Funding}
This research has received funding from the ERC program under Grant Agreement No. 810346 (Natural Bionics). The content of this publication is the sole responsibility of the authors. The European Commission or its services cannot be held responsible for any use that may be made of the information it contains.

\bibliographystyle{Frontiers-Harvard} 
\bibliography{biblio_SoftFoot_FrontRAI}

\end{document}